%% file: colm2026_conference.tex
\PassOptionsToPackage{table,dvipsnames}{xcolor}
\documentclass{article} 
\usepackage[final]{colm2026_conference}

\usepackage{microtype}
\usepackage{hyperref}
\usepackage{url}
\usepackage{booktabs}
\usepackage{graphicx}
\usepackage{subcaption}
\usepackage{tabularx}
\usepackage{multirow}
\usepackage{xcolor}
\usepackage{enumitem}
\usepackage{makecell}
\usepackage{adjustbox}
\usepackage{listings}
\usepackage{fontawesome5}
\usepackage[most]{tcolorbox}
\tcbuselibrary{raster}
\usepackage{siunitx}
\usepackage{lineno}
\usepackage{algorithm}
\usepackage{algorithmic}
\usepackage{wrapfig}
\usepackage{caption}
\usepackage{amsmath}
\usepackage{amssymb}
\usepackage{mathtools}
\usepackage{amsthm}
\usepackage{calc}
\usepackage[capitalize,noabbrev]{cleveref}

\definecolor{darkblue}{rgb}{0, 0, 0.5}
\hypersetup{
  colorlinks=true,
  citecolor=darkblue,
  linkcolor=darkblue,
  urlcolor=darkblue
}

\newcommand{\method}{InfMem}

\makeatletter
\pgfkeys{/tcb/raster valign/.style={}}
\makeatother

\lstdefinestyle{trace}{
  basicstyle=\ttfamily\footnotesize,
  breaklines=true,
  breakatwhitespace=true,
  columns=fullflexible,
  keepspaces=true,
  showstringspaces=false,
  frame=none,
  aboveskip=3pt,
  belowskip=3pt,
}

\newtcolorbox{bigcase}[1]{%
  enhanced,
  colback=gray!3,
  colframe=black!55,
  boxrule=0.9pt,
  arc=10pt,
  left=10pt,right=10pt,top=8pt,bottom=8pt,
  title=\textbf{#1},
  fonttitle=\bfseries,
}

\newtcolorbox{bigcaseblue}[1]{%
  enhanced,
  colback=blue!6,
  colframe=blue!70!black,
  boxrule=1.0pt,
  arc=10pt,
  left=10pt,right=10pt,top=8pt,bottom=8pt,
  title=\textbf{#1},
  fonttitle=\bfseries,
}

\newcommand{\cssec}[1]{\vspace{2pt}\noindent\textbf{#1}\par}

\theoremstyle{plain}

\theoremstyle{definition}

\theoremstyle{remark}

\title{InfMem: Learning System-2 Memory Control for Long-Context Agent}

\author{Xinyu Wang$^\spadesuit$\thanks{Equal contribution.}%
\quad Mingze Li$^\clubsuit$\footnotemark[1]%
\quad Peng Lu$^\heartsuit$%
\quad Xiao-Wen Chang$^\spadesuit$ \\
\bf Lifeng Shang$^\clubsuit$%
\quad Jinpeng Li$^\clubsuit$%
\quad Fei Mi$^\clubsuit$%
\quad Prasanna Parthasarathi$^\clubsuit$%
\quad Yufei Cui$^\clubsuit$\thanks{Corresponding authors: \texttt{xinyu.wang5@mail.mcgill.ca}, \texttt{yufei.cui@huawei.com}.} \\\\
\textsuperscript{$\spadesuit$} McGill University \
\textsuperscript{$\clubsuit$} Noah's Ark Lab\ \textsuperscript{$\heartsuit$} Universit\'e de Montr\'eal
}

\begin{document}

\ifcolmsubmission
\linenumbers
\fi

\maketitle

\input{sections/0_abstract}

\input{sections/1_introduction}

\input{sections/2_related_work}

\input{sections/3_framework}

\input{sections/4_training}
\input{sections/5_experiments}

\input{sections/6_main_result}
\input{sections/7_conclusion}
\input{sections/8_statement}

\bibliography{example_paper, colm2026_conference}
\bibliographystyle{colm2026_conference}

\appendix

\input{sections/X_appendix}

\end{document}

%% file: sections/0_abstract.tex
\begin{abstract}
Reasoning over ultra-long documents requires synthesizing sparse evidence scattered across distant segments under strict memory constraints.
While streaming agents enable scalable processing, their passive memory update strategy often fails to preserve low-salience \emph{bridging evidence} required for multi-hop reasoning.
We propose \textbf{InfMem}, a control-centric agent that instantiates System-2-style control via a \textsc{PreThink--Retrieve--Write} protocol.
InfMem actively monitors evidence sufficiency, performs targeted in-document retrieval, and applies evidence-aware joint compression to update a bounded memory.
To ensure reliable control, we introduce a practical SFT$\rightarrow$RL training recipe that aligns retrieval, writing, and stopping decisions with end-task correctness.
On ultra-long QA benchmarks from 32k to 1M tokens, InfMem consistently outperforms MemAgent across backbones.
Specifically, InfMem improves average absolute accuracy by \textbf{+10.17}, \textbf{+11.84}, and \textbf{+8.23} points on Qwen3-1.7B, Qwen3-4B, and Qwen2.5-7B, respectively, while reducing inference time by \textbf{3.9$\times$} on average (up to 5.1$\times$) via adaptive early stopping. \\Code is available at 
\href{https://github.com/UCMP13753/InfMem}{\faGithub\ \texttt{code}}.
\end{abstract}

%% file: sections/1_introduction.tex
\section{Introduction}
\label{sec:intro}

Large Language models (LLMs) have achieved remarkable performance across a wide range of tasks ~\citep{achiam2023gpt,touvron2023llama,deepseekdistillqwen_modelcard}. Despite these successes, they continue to exhibit limitations on tasks that require long-context abilities. In particular, long-document question answering increasingly demands inference over \emph{extreme-length} inputs--- ranging from multi-hundred-page legal contracts to codebases spanning hundreds of thousands of tokens---where correct answers hinge on a few \emph{sparse, widely scattered} facts rather than the global gist~\citep{shaham2022scrolls,an2024eval}. 
Existing long‑context LLM research largely focuses on improving the generic ability of models to operate over long inputs—either by (i) increasing the amount of text that can be ingested through architectural innovations~\citep{liu2023ring,gu2024mamba,yang2024parallelizing}  or (ii) encouraging models to generalize beyond their pre‑training context window~\citep{press2021train,su2024roformer,peng2024yarn}.  and reduce length‑extrapolation failures. While these methods mitigate the computation burden or preserve the basic language ability of LLM on extended inputs, the foundational LLMs still face challenges to make use of the relevant information in the extremely long contexts, as recent works show significant degradation when models must access correct information in the middle of long inputs~\citep {lost_in_the_middle,lost_in_the_middle_ir}. 

Instead of feeding an entire ultra‑long document into an LLM directly, a practical alternative is streaming processing: the document is segmented into chunks read sequentially, and the model maintains a fixed‑size bounded memory that is selectively overwritten at each step, yielding constant memory cost and linear‑time computation~\citep{packer2023memgpt,yu2025memagent}. While these methods improve efficiency, they introduce an unidiretional optimization bias. Specifically, every chunk is forced through the same fixed‑size memory bottleneck, regardless of the complexity of the underlying question. This uniform treatment implicitly assumes that all segments of the document contribute equally to the final answer. However, long‑document QA naturally contains a mixture of \emph{high‑complexity} and \emph{low‑complexity} queries. Some questions require multi‑step reasoning over subtle cross‑section dependencies, while others only demand direct retrieval of a single explicit fact. This mismatch creates a structural dilemma: to preserve information for complex questions, the memory must retain rich, high‑fidelity representations of earlier content; yet to remain within the fixed memory budget, the system must heavily compress each chunk, inevitably discarding fine‑grained details.

This structural dilemma exposes a deeper limitation of streaming-based
long-context processing: the system cannot adapt its reasoning depth or memory
preservation to the complexity of the question. Complex queries demand
multi-step abstraction and the retention of subtle cross-section dependencies,
yet a fixed-size memory with uniform compression treats all incoming
information homogeneously. In the language of dual-process theory
\citep{kahneman2011thinking}, such architectures are locked into fast,
reflexive "System-1" heuristics where deliberate, state-dependent System-2
control is needed. While recent work on test-time scaling
\citep{wei2022chain,jaech2024openai,deepseekai2025deepseekr1} shows that LLMs
can dynamically allocate computation at inference time, no existing streaming
framework channels this capacity into principled memory control for
long-context QA.

We address this gap with InfMem, a bounded-memory agent that replaces passive
compression with an explicit, System-2-style control loop
\citep{kahneman2011thinking, sumers2023cognitive}, instantiated as a
\textsc{PreThink--Retrieve--Write} protocol. At each streaming step,
\textsc{PreThink} assesses whether the accumulated evidence suffices for the
query; if not, \textsc{Retrieve} issues targeted queries against a global
document index, and \textsc{Write} jointly compresses the retrieved passages,
the current chunk, and the existing memory within a fixed budget. Once evidence
is deemed sufficient, the loop stops early, protecting consolidated memory from
redundant overwrites and curtailing computation. To make these control
decisions reliable rather than heuristic, we combine supervised fine-tuning on
reasoning-correct trajectories with reinforcement learning that aligns
retrieval, writing, and stopping with end-task correctness and efficiency.
On ultra-long QA benchmarks (up to 1M tokens), InfMem outperforms the state of the art by over \textbf{10 points} across three Qwen-series backbones while reducing inference latency by \textbf{3.9$\times$} on average via adaptive early stopping.

%% file: sections/2_related_work.tex
\section{Related Work}
\label{sec:related_work}
\paragraph{Long-Context Modeling and Efficiency.}
Frontier models now scale to million-token regimes~\citep{qwen2025qwen3next,wan2025qwenlongl1,yang2025qwen25_1m,alibabacloud2025qwenlong}, and efficient architectures such as linear attention and SSMs reduce the quadratic cost of self-attention~\citep{gu2024mamba,yang2024parallelizing}. However, both transformers and their efficient alternatives consistently suffer from length generalization issues when trained on short sequences and tested on long ones, requiring careful intervention either before or after training~\citep{train_short_test_long,yarn,mamba_modulation}. Limited capacity is a general challenge when processing long contexts for architectures with recurrent states or local windows, and incorporating additional modules to augment this capacity can improve performance on retrieval-heavy tasks~\citep{wang2025resona,match}. Yet expanded capacity alone does not ensure effective reasoning: retrieval-heavy tasks still exhibit "lost-in-the-middle" degradation~\citep{liu2024lost,weston2023system}, and monolithic processing offers no explicit control over which evidence is retained or discarded. Integrating retrieval natively into the reasoning chain improves context fidelity~\citep{wang2025care}, but presumes the document fits within the context window and offers no mechanism for streaming consolidation under a memory budget.
 
\paragraph{Bounded-Memory Agents.}
\citet{zhang2023large} formulate LLMs as semi-parametric RL agents that learn to retrieve and update memory.
MEM1~\citep{zhou2025mem1}, Memory-R1~\citep{yan2025memory}, and MemGPT~\citep{packer2023memgpt} introduce dedicated memory mechanisms but primarily target conversational settings~\citep{maharana2024evaluating}, prioritizing state tracking or persona consistency rather than evidence aggregation over a single document.
Beyond textual memory, context can also be consolidated \emph{parametrically}---by transforming it into weight updates~\citep{cao2025infiniteicl} or via lifelong model editing~\citep{wang2026rlsedit}---which faces an analogous plasticity--stability dilemma between absorbing new evidence and preserving consolidated knowledge.
For long-document QA, MemAgent~\citep{yu2025memagent} is the most directly relevant: it streams chunks sequentially and recurrently compresses each chunk with the existing memory. However, this fixed routine cannot assess memory sufficiency, revisit lost context, or terminate early.

\begin{figure*}[t]
  \centering
  \includegraphics[width=0.85\textwidth]{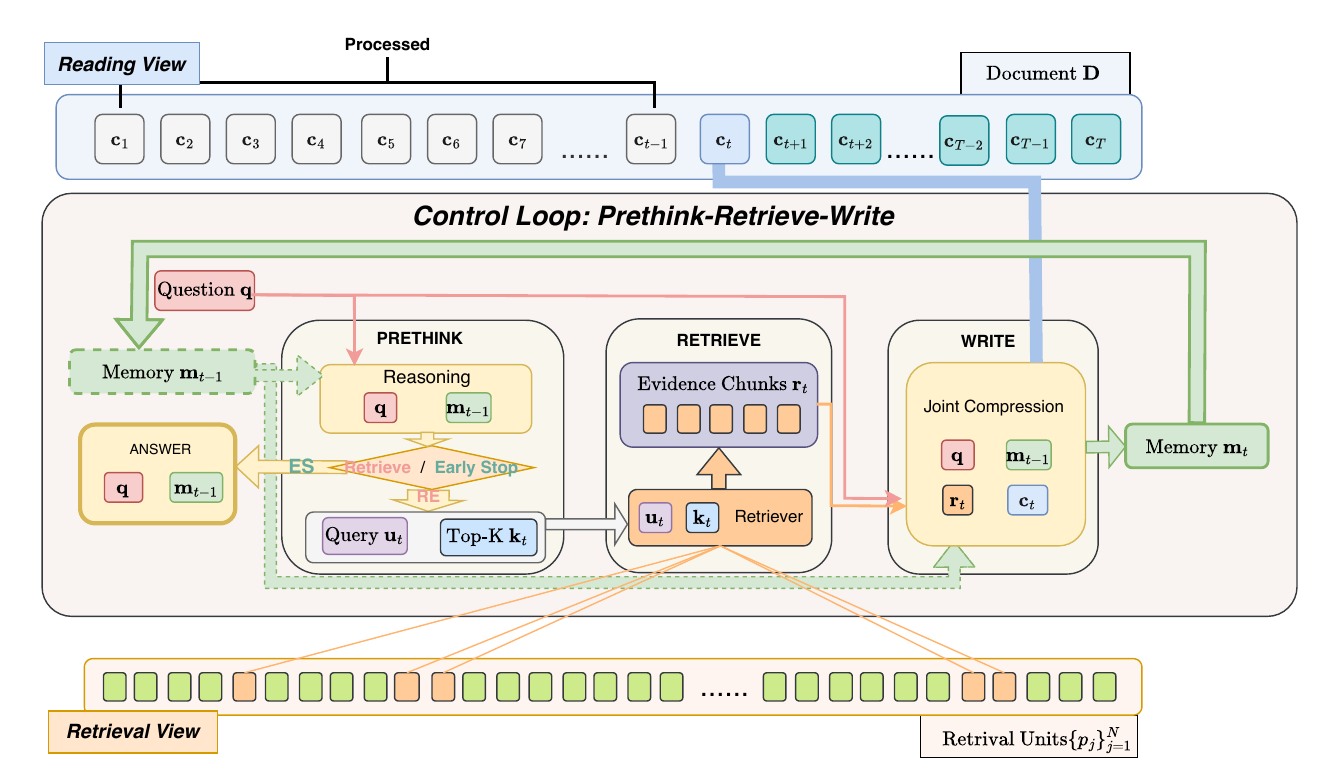}
\caption{InfMem runs a \textbf{System-2 control loop} (\textsc{PreThink}--\textsc{Retrieve}--\textsc{Write}) over bounded memory. \textsc{PreThink} monitors sufficiency and triggers early stops; \textsc{Retrieve} fetches sparse evidence to bridge logical gaps; \textsc{Write} compresses evidence with the current stream to update memory, preserving fidelity under extreme context lengths.}
\label{fig:infmem_fw}
\label{fig:infmem_fw}
  \label{fig:infmem_fw}
\end{figure*}

%% file: sections/3_framework.tex
\section{InfMem: Streaming Inference  with a Control Loop}

In this section, we begin by formally defining the streaming long-context reasoning problem, then introduce InfMem, our inference protocol that augments such agents with a closed-loop control mechanism to dynamically manage what to retain, retrieve, or discard under a constrained memory budget.
\label{sec:framework}
\subsection{Problem Formulation}
\label{subsec:setup}
Given a question $q$ and a document $D$ too long for a single context window, the goal is to produce answer $\hat{y}$ from evidence distributed throughout $D$. Following \citet{yu2025memagent}, we segment $D$ into $T$ ordered chunks $\{c_t\}_{t=1}^{T}$ processed left-to-right, with the agent maintaining a bounded memory $m_t$ under a fixed token budget $|m_t| \leq M$. After each chunk, the agent overwrites one memory entry to form $m_{t+1}$, keeping per-step context constant and end-to-end cost linear in $T$.

\subsection{Control Loop: PreThink–Retrieve–Write with Early Stop}
\label{subsec:workflow}
\begin{minipage}{0.45\textwidth}
\begin{algorithm}[H]
\small
\caption{InfMem Control Loop}
\label{alg:infmem}
\begin{algorithmic}[1]
\REQUIRE question $q$;\; streaming chunks $\{c_t\}_{t=1}^T$;\; retrieval units $\{p_j\}_{j=1}^N$;\; budget $M$
\STATE $m_0 \leftarrow \varnothing$
\FOR{$t = 1$ \textbf{to} $T$}
    \STATE $(a_t,\, u_t,\, k_t) \leftarrow \textsc{PreThink}(q,\, m_{t-1})$
    \IF{$a_t = \textsc{Stop}$}
        \STATE \textbf{break}
    \ENDIF
    \STATE $r_t \leftarrow \textsc{Retrieve}(u_t,\, k_t;\, \{p_j\})$
    \STATE $m_t \leftarrow \textsc{Write}(q,\, m_{t-1},\, c_t,\, r_t;\, M)$ \quad s.t.\ $|m_t| \leq M$
\ENDFOR
\STATE $\hat{y} \leftarrow \textsc{Answer}(q,\, m_\star)$
\STATE \textbf{return} $\hat{y}$
\end{algorithmic}
\end{algorithm}
\end{minipage}
\hfill
\begin{minipage}{0.53\textwidth}
\paragraph{Overall Pipeline.} \textbf{InfMem} boosts models with a control loop.
At step $t$, it maintains a bounded memory $m_{t-1}$ as its intermediate state and proceeds via three operations.
First, \textsc{PreThink}, conditioned solely on $(q, m_{t-1})$, \textbf{monitors} whether the accumulated evidence is sufficient to answer $q$.
If so, the agent emits \textsc{Stop} and terminates early, avoiding unnecessary computation.
Otherwise, it emits \textsc{Retrieve} together with a synthesized retrieval query and a predicted fetch count $k_t$, and invokes \textsc{Retrieve} to \textbf{seek} sparse evidence globally from the source document, producing a compact retrieved context $r_t$.
Finally, \textsc{Write} \textbf{updates} the memory by jointly compressing the incoming chunk $c_t$, the retrieved context $r_t$, and the prior memory $m_{t-1}$ under the fixed budget $M$, yielding $m_t$ with $|m_t| \leq M$.
We detail each operation below (see also Figure~\ref{fig:infmem_fw}).
\end{minipage}

\paragraph{\textsc{PreThink}.}
\textsc{PreThink} is a state-dependent controller. Given $(q, m_{t-1})$, it outputs a structured control record $c_t = (a_t, u_t, k_t)$ that specifies the step-$t$ action:

\begin{itemize}[leftmargin=*, itemsep=2pt, topsep=2pt]
    \item \textsc{Action} $a_t \in \{\textsc{``STOP''}, \textsc{``SEARCH''}\}$: 
    whether the current memory is sufficient to answer $q$ (stop) or additional in-document evidence is needed (retrieve);
    
    \item \textsc{Query} $u_t$ (if $a_t=\textsc{``SEARCH''}$): 
    a single dynamic query synthesized from $(q, m_{t-1})$;
    
    \item \textsc{TopK} $k_t \in \{1,\ldots,K_{\max}\}$ (if $a_t=\textsc{``SEARCH''}$): 
    the number of retrieval units to fetch.
\end{itemize}
Together, $(a_t, u_t, k_t)$ define the control decisions at step $t$:
\emph{whether to stop}, and if continuing, \emph{what to search} and \emph{how much to search}.
Optionally, \textsc{PreThink} may also emit a brief natural-language rationale (e.g., missing evidence or subgoals) to improve interpretability
and prompting, but these auxiliary fields do not affect execution beyond the induced $u_t$.

\paragraph{\textsc{Retrieve}.}
While the document is processed sequentially as coarse streaming chunks, we strictly distinguish the \emph{reading view} from the \emph{retrieval view}.
We pre-construct a finer-grained set of \emph{retrieval units} $\{p_j\}_{j=1}^N$ (e.g., paragraphs) from the same document.
Unlike the coarse streaming chunks, these units are compact and globally indexed. When triggered by \textsc{PreThink}, \textbf{InfMem} can search any part of the document (past or future) to retrieve the top-$k_t$ units and summarize them into a concise context $r_t$, while preserving the coarse-grained reading flow.
If $a_t=\textsc{``SEARCH''}$, \textbf{InfMem} retrieves top-$k_t$ relevant retrieval units from the same document (no external corpus) and concatenates
them into a compact context:
\begin{equation}
\begin{split}
    P_t &\leftarrow \textsc{Retrieve}(u_t, k_t; \{p_1,\ldots,p_N\}), \\
    r_t &\leftarrow \mathrm{Concat}(P_t),
\end{split}
\label{eq:retrieve}
\end{equation}
with separators and (optionally) unit identifiers to preserve provenance.

\paragraph{\textsc{Write}.}
If $a_t=\textsc{``SEARCH''}$, \textbf{InfMem} overwrites the memory with a bounded new state:
\begin{equation}
    m_t \leftarrow \textsc{Write}(q, m_{t-1}, c_t, r_t; M),
    \quad \text{s.t. } |m_t|\le M.
\label{eq:write}
\end{equation}
\textsc{Write} has access to $(q, m_{t-1}, c_t)$ as well as the retrieved evidence $r_t$, and then performs evidence-aware composition:
it connects the retrieved support $r_t$ with the newly observed content in $c_t$ in order to identify and encode the composition-critical facts and the bridging links
into a bounded updated memory.
We refer this overwrite update to as \textbf{joint compression}, 
where the retrieval is used \emph{for writing} to shape the memory update.

\paragraph{\textsc{Early Stop}.}
If $a_t=\textsc{``STOP''}$ at a step, 
the agent halts the retrieval and the memory updates: it directly produces the final answer using the current memory.
Otherwise, it continues until the end of the chunk stream ($t=T$).
After termination (early stopping or reaching the end-of-sequence), \textbf{InfMem} generates:
\[
\hat{y} \leftarrow \textsc{Answer}(q, m_\star),
\]
where $m_\star$ denotes the final memory state at termination.

%% file: sections/4_training.tex
\section{Training InfMem}
The \textsc{PreThink--Retrieve--Write} protocol poses training challenges absent in standard language modeling: the model must produce heterogeneous outputs (retrieval decisions, free-form queries, and compressed memory) whose quality is only revealed at the end of a long rollout, with no ground-truth labels for intermediate decisions. This motivates a two-stage recipe. \textbf{Stage~1 (SFT)} teaches the model to produce protocol-valid outputs---well-formed function calls, schema-consistent memory updates, and formatted answers. \textbf{Stage~2 (RL)} then optimizes the control policy---when to stop, what to retrieve, and how to compress---via outcome-based rewards propagated through the full trajectory.

\subsection{SFT Warmup via Supervised Distillation}
\label{subsec:warmup}
 
We distill a strong teacher (\texttt{Qwen3-32B}) into a smaller student using trajectories from the \textsc{PreThink--Retrieve--Write} loop. For each question, the teacher executes the full protocol, optionally guided by golden question decompositions (e.g., for MuSiQue) to elicit cleaner retrieval traces, in the spirit of evidence-grounded trajectory synthesis~\citep{li2026evipath}. Crucially, these hints are given \emph{only to the teacher}; the student never observes decompositions, gold document IDs, or any teacher-side annotations. The subsequent RL stage operates entirely without such hints yet yields substantial additional gains (Table~\ref{tab:lb_ruler_gains_transposed}), confirming that the learned control policy---not teacher supervision---drives the final performance.
\paragraph{Data filtering and training objective.}
We construct a warmup set from HotpotQA, SQuAD, and MuSiQue (Appendix~\ref{app:sft_data}).
Only trajectories whose final answer is correct ($\mathrm{EM}(\hat{y},y){=}1$) are retained; string and regex filters remove any ground-truth leakage.
Each retained rollout is serialized into a protocol-formatted dialogue $\tau$, and the student is trained with masked next-token prediction on \emph{agent response tokens} only (masking all system/user/prompt tokens):
\begin{equation}
\mathcal{L}_{\text{SFT}}
=
-\sum_{\tau\in\mathcal{D}_{\text{SFT}}}
\sum_{i \in \mathcal{Y}(\tau)}
\log \pi_\theta\!\big(y_i \mid \mathrm{prefix}_i(\tau)\big),
\label{eq:sft}
\end{equation}
where $\mathcal{Y}(\tau)$ indexes the response tokens and $\mathrm{prefix}_i(\tau)$ denotes all preceding tokens.
Gradients propagate through all realized steps, jointly supervising the control records, bounded memory updates, and final answers.
In practice, SFT primarily teaches the \emph{mechanics} of the protocol; the control policy is refined in the subsequent RL stage (\S\ref{subsec:rl}).
 
\subsection{RL Alignment with Reward Design}
\label{subsec:rl}

While SFT ensures protocol-compliant execution, it does not optimize the \emph{control policy} itself---when to stop, what to retrieve, and how to compress under the overwrite budget.
We apply RL to align these decisions with end-task success and efficiency.
A central design challenge is that InfMem's control loop produces long-horizon trajectories with \emph{delayed, sparse} feedback: only the final answer is verifiable, yet the decisive actions---a well-targeted retrieval query, a faithful memory compression, or a timely stop---may occur many steps earlier.
Our reward design addresses this by combining four complementary signals that together supervise task correctness, stopping efficiency, and protocol integrity.

\paragraph{Task correctness reward.}
The primary signal is a binary ground-truth reward on the final predicted answer:
\begin{equation}
R_{\text{gt}}(\hat{y},y)
=
\mathbf{1}\{\mathrm{equiv}(\hat{y},y)\},
\label{eq:gt_reward}
\end{equation}
where $\mathrm{equiv}(\cdot,\cdot)$ follows the official benchmark evaluation protocol (e.g., exact-match normalization).
This reward drives the agent to retrieve the right evidence and compress it faithfully, but alone provides no signal for \emph{when} to stop.

\paragraph{Early-stop shaping.}
Continuing after the memory already suffices wastes compute and risks displacing consolidated evidence via redundant overwrites.
We introduce a shaping term that rewards stopping soon after the memory first becomes sufficient.
Let $t_{\text{first}}$ be the earliest step at which the current memory alone yields a correct answer (EM=1 under the official evaluation protocol), and let $t_{\text{stop}}$ be the agent's actual stopping step.
With $d = t_{\text{stop}} - t_{\text{first}}$:
\begin{equation}
R_{\text{early}}=\gamma^{\,d-1}, \quad \gamma\in(0,1),
\label{eq:early_stop_reward}
\end{equation}
so $R_{\text{early}}=1$ for immediate stopping after sufficiency, decaying exponentially with delay.
The sufficiency check uses the same EM protocol as final evaluation; while it may occasionally misjudge on hard multi-hop cases, the group-level advantage normalization during optimization averages over $G$ rollouts, smoothing individual noisy signals.

\paragraph{Protocol-soundness verifiers.}
Long-horizon agentic rollouts are fragile: a single malformed function call or truncated memory update can corrupt all downstream steps.
We add two binary rollout-level verifiers to keep RL exploration within protocol-valid regions:
(i)~$R_{\text{call}} = 1$ iff all function calls are well-formed and parsable;
(ii)~$R_{\text{mem}} = 1$ iff every memory-update step outputs a complete, non-truncated \textsc{UpdatedMemory} field within the fixed budget.
The four signals are combined into a single outcome reward per rollout:
\begin{equation}
R = \sum_{w} \alpha_{w} R_w, \quad w \in \{\text{gt, early, call, mem}\},
\label{eq:final_reward}
\end{equation}
where all coefficients $\alpha_w$ are specified in \S\ref{subsec:setup}.

\paragraph{Optimization.}
We optimize the composite reward $R$ using group relative 
policy optimization (GRPO). Each rollout spans multiple 
memory-update steps and a final answering step, with $R$ 
shared across all preceding steps for long-horizon credit 
assignment. For each query we sample $G$ rollouts, compute 
mean-centered advantages $\hat{A}_i = R_i - \bar{R}$, and 
optimize:
\begin{equation}
J(\theta) = \mathbb{E}_{i,t}\!\Big[
\min\!\big(r_{i,t}(\theta)\hat{A}_i,\; 
\mathrm{clip}(r_{i,t}(\theta),1{-}\epsilon,1{+}\epsilon)
\hat{A}_i\big)
- \beta D_{\mathrm{KL}}\!\big(
\pi_\theta \| \pi_{\mathrm{ref}}\big)
\Big],
\label{eq:rl_obj}
\end{equation}
where $r_{i,t}(\theta) = \pi_\theta(a_{i,t} \mid s_{i,t}) 
/ \pi_{\theta_{\text{old}}}(a_{i,t} \mid s_{i,t})$.

%% file: sections/5_experiments.tex
\section{Experiments and Results}
\subsection{Experimental Setup}
\label{sec:datasets}

\paragraph{Datasets and backbones.} We utilize four datasets spanning a spectrum of reasoning demands: \texttt{SQuAD}~\citep{squad_dataset}, \texttt{HotpotQA}~\citep{hotpot_qa}, \texttt{2WikiMultiHopQA}~\citep{2wiki}, and \texttt{MuSiQue}~\citep{musique}.
These corpora form the basis for constructing our synthetic long-context training data and evaluation benchmarks, as detailed in \S\ref{app:data}. In addition, we report results on \texttt{LongBench}~\citep{DBLP:conf/acl/BaiLZL0HDLZHDTL24}, a standardized long-context benchmark suite that evaluates LLMs under unified prompts and consistent scoring across diverse long-document QA tasks as complements to our controlled settings. We evaluate InfMem on \texttt{Qwen3-1.7B}, \texttt{Qwen3-4B}~\citep{DBLP:journals/corr/abs-2505-09388}, and \texttt{Qwen2.5-7B-Instruct}~\citep{DBLP:journals/corr/abs-2412-15115} as base policies $\pi_\theta$ for both SFT and RL stages.

\label{subsec:setup}

\paragraph{Training Data and Settings.} We pack synthetic \textsc{PreThink--Retrieve--Write} trajectories from \texttt{HotpotQA}, \texttt{SQuAD}, and \texttt{MuSiQue} (\S\ref{app:sft_data}) into \texttt{32k}-token sequences for SFT, using \texttt{Qwen3-32B} as the distillation teacher. RL training uses a long-context variant of \texttt{HotpotQA} (\S\ref{app:RL_data}) for dense multi-hop signals. For Stage~1 (SFT), we use a cosine-decayed learning rate of $4.0\times10^{-5}$ with batch size 256; \texttt{Qwen3-1.7B} and \texttt{4B} are trained for \texttt{1 epoch}, while \texttt{Qwen2.5-7B-Instruct} is trained for {4 epochs}. For Stage~2 (RL), we apply GRPO ($G{=}4$ rollouts, temperature $1.0$, top-$p{=}1.0$, $\beta{=}0.001$) with a constant learning rate of $1\times10^{-6}$ and batch size 128 (mini-batch 8). Detailed hyperparameters are provided in \S\ref{subsec:setup}.

\label{sec:experiments}

%% file: sections/6_main_result.tex
\vspace{-0.1cm}
\label{sec:main_results}

\input{tables/table_diff_scale}

\subsection{Results on Various Input Lengths up to 1M Tokens}
Table \ref{tab:generality_backbones} reports results on synthesized long-context QA benchmarks evaluated across \texttt{Qwen3-4B} and \texttt{Qwen2.5-7B}. We compare our methods against standard long-context baselines (YaRN, RAG) and strong iterative or memory-construction frameworks like RESP and Mem-$\alpha$ (see Appendix~\ref{app:baseline_details} for full reproduction details). Two further retrieval-based baselines evaluated under a matched in-document corpus---Search-R1~\citep{jin2025searchr1}, an RL-trained retrieval-decision policy, and Self-RAG~\citep{asai2024self}, in both a prompted train-free variant and a checkpoint we train ourselves---are reported in Appendix~\ref{app:extra_baselines}; InfMem outperforms both even when they use substantially larger backbones.

Models equipped with the length generalization method (YaRN~\citep{peng2024yarn}) or conventional retrieval mechanism (RAG~\citep{DBLP:conf/nips/LewisPPPKGKLYR020}) degrade sharply beyond 128k tokens, often collapsing to single digits at 1M. iterative RAG system (RESP~\citep{jiang2025retrievesummarizeplanadvancing}) and recent bounded memory agent method (Mem-$\alpha$~\citep{wang2025mem}) mitigate this decline but still underperform both MemAgent and InfMem on average. InfMem consistently achieves the strongest results across all settings.

This performance advantage becomes particularly pronounced on complex multi-hop reasoning tasks where evidence is scattered. For instance, on the MuSiQue benchmark at 1M tokens (using Qwen3-4B+RL), InfMem achieves a score of 56.86 compared to MemAgent~\citep{yu2025memagent}'s 35.91. Similarly, on 2WikiMultiHopQA at 1M tokens, InfMem reaches 66.39 versus 35.18. On single-hop tasks like SQuAD, the performance gap narrows slightly, as MemAgent's fixed, reactive compression is generally sufficient for direct fact extraction. However, for multi-hop queries, MemAgent's recurrent architecture is highly susceptible to gradual information loss over extended horizons. In contrast, InfMem leverages explicit reasoning and decision-making capabilities to dynamically decompose complex questions into manageable sub-problems. Finally, the proposed SFT$\to$RL training recipe further amplifies these structural advantages by optimizing the agent's sequential decision-making and adaptive stopping policies. Consequently, RL-InfMem establishes a decisive lead, outperforming RL-MemAgent by an average margin of over 10 points across both evaluated backbones.

\subsection{Results on Standard LongBench}

\input{tables/longbench_task_T}

Crucially, these gains are not confined to our synthesized ultra-long setting. As shown in Table \ref{tab:longbench_delta_table}, the performance improvements transfer to LongBench QA, where contexts are shorter but information density is higher—placing greater emphasis on evidence analysis and selection rather than on maintaining memory over long horizons (see detailed discussion in Section \ref{app:quality}).

Across backbone architectures, InfMem consistently outperforms MemAgent in both train-free and RL-enhanced configurations. Notably, while InfMem already leads without reinforcement learning, incorporating RL further widens the gap over strong baselines such as YaRN.
Overall, these results indicate that InfMem strengthens not only robustness at extreme lengths (up to 1M tokens) but also the quality of reasoning-oriented evidence management on standard long-context QA benchmarks, demonstrating versatility across diverse task conditions.

\section{Analysis and Discussion}
\begin{figure*}[t!]
    \centering
    \includegraphics[width=.95\textwidth]{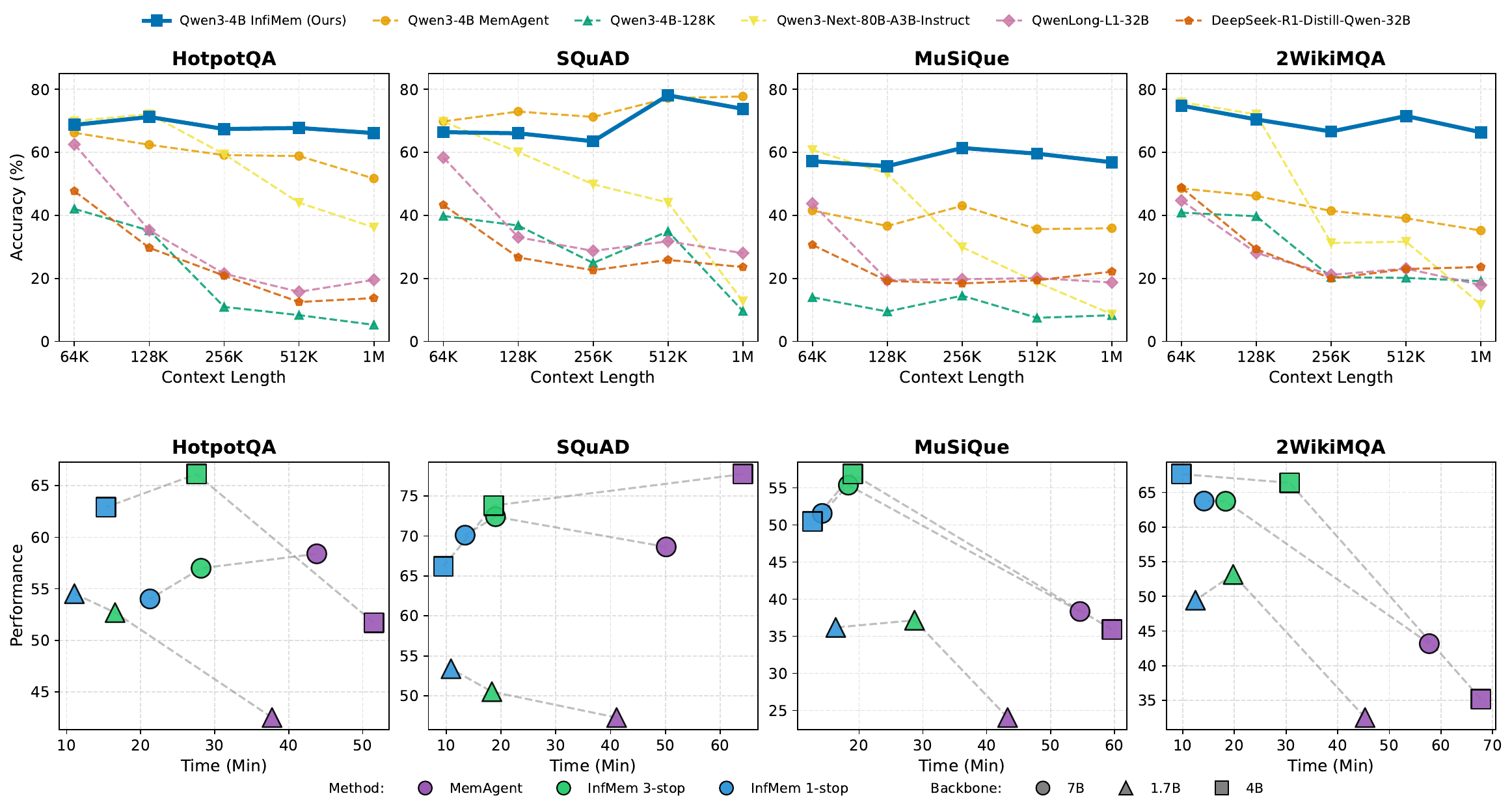}
    \caption{\textbf{Robustness and Inference Efficiency of InfMem on Synthesized QA Benchmarks.} 
\textbf{Top row:} Performance scaling of \texttt{Qwen3-4B} across context lengths from 64K to 1M tokens. InfMem demonstrates remarkable robustness, maintaining consistent accuracy without performance degradation as context length increases. 
\textbf{Bottom row:} Inference efficiency versus QA performance at 1M context length. InfMem exhibits exceptional proficiency in long-range multi-hop reasoning, preserving high-fidelity performance without the computational overhead typically associated with extreme sequence lengths.}
    \label{fig:4b_trend}
\end{figure*}
\subsection{Robustness Analysis across Context Length}
\label{subsec:scaling_4b}

We further compare our method with strong LLMs supporting long-context, including \texttt{Qwen3-Next-80B-A3B-instruct}~\citep{qwen2025qwen3next}, \texttt{QwenLong-L1-32B}~\citep{alibabacloud2025qwenlong} and \texttt{Deepseek-R1-Distill-Qwen-32B}~\citep{deepseekdistillqwen_modelcard}.
The top rows of Figure~\ref{fig:4b_trend} summarizes long-context scaling on \texttt{Qwen3-4B} up to 1M tokens. Despite extended context windows, accuracy often deteriorates in the ultra-long regime where evidence is sparse and separated by long gaps. InfMem remains substantially more stable beyond 128K tokens, and its advantage grows with length, especially on multi-hop datasets. We attribute this to sufficiency-aware control over retrieval and memory writing, which mitigates long-horizon drift from repeated compression and enables targeted recovery of missing bridging facts before updating memory. Qualitative case studies are provided in Section~\ref{app:case}.

\subsection{Inference Efficiency Analysis}
\label{app:early_stop}

\input{tables/token_cost}
The bottom row of Figure~\ref{fig:4b_trend} illustrates the efficiency-quality trade-off on 1M-token tasks. Across \texttt{Qwen3-4B} and \texttt{Qwen2.5-7B}, InfMem outperforms MemAgent on both axes: it improves accuracy by 11.80, 11.67, and 7.73 points, while reducing latency by 5.1x, 3.3x, and 3.3x. The conservative 3-stop policy further gains 2.76 points yet remains under half the runtime of MemAgent. These results confirm that InfMem reliably stops upon collecting sufficient evidence, avoiding redundant steps and establishing a superior efficiency-accuracy frontier.  
A more detailed analysis of early stopping behaviors and their direct impact on latency is provided in Appendix~\ref{appendix:early_stop}.

\paragraph{Token cost.} Since wall-clock time is partly shaped by serving-level batching, we additionally report the decode cost of each method. Table~\ref{tab:token_cost} lists generated tokens per query at $\approx$112k and $\approx$896k context. Among the methods that remain functional at ultra-long inputs, InfMem is the cheapest: at $\approx$896k it decodes 33.1K tokens per query against 117.4K for MemAgent ($0.28\times$) and 115.0K for Mem-$\alpha$, and its cost grows only $\approx5\times$ from 112k to 896k while MemAgent's grows $\approx11\times$. A per-length breakdown of tokens and latency, together with retrieval and index-storage accounting, is given in Appendix~\ref{app:cost_accounting}.

\subsection{Further Analysis and Ablation Study}
Beyond the main results, we provide comprehensive ablation and further studies in Appendix \ref{app:ablation}. These include retrieval chunk size selection, ablation on the thinking mode, and an analysis of memory retention. Appendix~\ref{app:extended} additionally reports a zero-shot transfer of the InfMem protocol to a non-Qwen backbone (\S\ref{app:cross_family}), where the marginal gain on multi-hop QA is nearly identical to the one observed on Qwen3-4B.

%% file: tables/table_diff_scale.tex
\begin{table*}[t]
\centering
\caption{Performance on synthesized RULER-style benchmarks across \texttt{Qwen3-4B} and \texttt{Qwen2.5-7B} under increasing context lengths. To ensure robust baselines, RESP directly uses \texttt{Qwen3-32B}, while Mem-$\alpha$ pairs its context manager with \texttt{Qwen3-32B} for question answering. Overall, MemAgent and InfMem provide consistent train-free gains over baselines (including YaRN and RAG), with RL further amplifying these improvements.}
\scriptsize
\setlength{\tabcolsep}{2.2pt} 
\renewcommand{\arraystretch}{1.12}
\resizebox{\textwidth}{!}{%
\begin{tabular}{
l|
cc|
ccc>{\columncolor{gray!15}}c|c>{\columncolor{gray!30}}c|
ccc>{\columncolor{gray!15}}c|c>{\columncolor{gray!30}}c
}
\toprule
\multirow{3}{*}{\textbf{Metric}} &
\multicolumn{2}{c|}{\textbf{Other approach}} &
\multicolumn{6}{c|}{\textbf{Qwen3-4B}} &
\multicolumn{6}{c}{\textbf{Qwen2.5-7B}} \\
\cmidrule(lr){2-3}\cmidrule(lr){4-9}\cmidrule(lr){10-15}
&
\multirow{2}{*}{\textbf{RESP}} & \multirow{2}{*}{\textbf{mem-$\alpha$}} & 
\multicolumn{4}{c|}{\textbf{Framework}} & \multicolumn{2}{c|}{\textbf{+RL}} &
\multicolumn{4}{c|}{\textbf{Framework}} & \multicolumn{2}{c}{\textbf{+RL}} \\
\cmidrule(lr){4-7}\cmidrule(lr){8-9}\cmidrule(lr){10-13}\cmidrule(lr){14-15}
&
& &
\textbf{YaRN} & \textbf{RAG} & \textbf{MemAgent} & \textbf{InfMem} & \textbf{MemAgent} & \textbf{InfMem} &
\textbf{YaRN} & \textbf{RAG} & \textbf{MemAgent} & \textbf{InfMem} & \textbf{MemAgent} & \textbf{InfMem} \\
\midrule

\textbf{avg} &
30.29 & 37.42 &
25.45 & 26.05 & \underline{43.61} & \textbf{50.25} & 54.56 & \textbf{66.40} &
21.41 & 19.77 & \underline{37.06} & \textbf{47.73} & 52.07 & \textbf{60.30} \\

\midrule
\multicolumn{15}{l}{\textbf{HQA}} \\
28k  &
44.22 & 46.18 &
50.77 & 48.46 & \underline{52.55} & \textbf{59.73} & 71.18 & \textbf{71.44} &
35.70 & 33.51 & \underline{44.96} & \textbf{45.70} & \textbf{65.58} & 59.20 \\
56k  &
45.59 & 42.42 &
42.07 & 43.69 & \underline{51.27} & \textbf{58.69} & 66.21 & \textbf{68.73} &
31.74 & 29.87 & \underline{45.93} & \textbf{48.56} & \textbf{62.88} & 62.23 \\
112k &
41.10 & 45.62 &
35.19 & 42.52 & \underline{44.02} & \textbf{51.33} & 62.42 & \textbf{71.24} &
25.42 & 31.45 & \underline{42.76} & \textbf{47.98} & \textbf{61.55} & 57.75 \\
224k &
25.80 & 40.02 &
10.96 & 24.11 & \underline{44.68} & \textbf{47.82} & 59.12 & \textbf{67.42} &
13.93 & 16.94 & \underline{34.77} & \textbf{49.65} & 59.95 & \textbf{60.55} \\
448k &
20.43 & 35.99 &
8.34 & 13.27 & \underline{40.47} & \textbf{51.71} & 58.84 & \textbf{67.75} &
9.23 & 8.61 & \underline{33.07} & \textbf{46.70} & 57.09 & \textbf{62.34} \\
896k &
16.72 & 35.84 &
5.26 & 3.73 & \underline{40.03} & \textbf{49.07} & 51.70 & \textbf{66.13} &
3.91 & 2.39 & \underline{34.47} & \textbf{42.60} & \textbf{58.39} & 57.51 \\

\midrule
\multicolumn{15}{l}{\textbf{SQuAD}} \\
32k  &
46.08 & 45.89 &
48.70 & \underline{55.66} & 53.82 & \textbf{65.75} & \textbf{69.49} & 65.31 &
34.36 & 36.80 & \underline{45.02} & \textbf{55.77} & 61.95 & \textbf{61.70} \\
64k  &
48.29 & 45.36 &
39.80 & 49.91 & \underline{54.73} & \textbf{61.07} & \textbf{69.84} & 66.42 &
31.55 & 33.55 & \underline{47.06} & \textbf{53.98} & 57.94 & \textbf{64.19} \\
128k &
40.36 & 45.10 &
36.79 & 45.05 & \underline{51.80} & \textbf{64.17} & \textbf{72.96} & 66.05 &
29.44 & 27.93 & \underline{49.15} & \textbf{54.23} & 58.26 & \textbf{58.82} \\
256k &
38.03 & 40.86 &
24.89 & 35.76 & \underline{46.23} & \textbf{59.23} & \textbf{71.24} & 63.53 &
27.50 & 22.15 & \underline{41.83} & \textbf{53.03} & 53.23 & \textbf{61.71} \\
512k &
31.62 & 52.68 &
34.93 & 26.36 & \underline{51.62} & \textbf{64.99} & 77.21 & \textbf{78.12} &
20.23 & 20.70 & \underline{50.92} & \textbf{63.08} & \textbf{69.85} & 69.27 \\
1M   &
17.50 & 49.72 &
9.63 & 5.26 & \underline{48.91} & \textbf{59.38} & \textbf{77.74} & 73.81 &
4.59 & 2.57 & \underline{44.97} & \textbf{55.99} & \textbf{68.63} & 67.71 \\

\midrule
\multicolumn{15}{l}{\textbf{MuSiQue}} \\
32k  &
28.45 & 29.17 &
19.65 & 19.84 & \underline{29.02} & \textbf{41.35} & 41.79 & \textbf{56.58} &
19.93 & 18.37 & \underline{31.45} & \textbf{37.09} & 36.67 & \textbf{46.27} \\
64k  &
31.52 & 25.33 &
14.00 & 11.94 & \underline{34.03} & \textbf{38.06} & 41.55 & \textbf{57.19} &
17.27 & 14.13 & \underline{26.29} & \textbf{36.73} & 32.82 & \textbf{46.05} \\
128k &
26.78 & 23.08 &
9.48 & 15.78 & \underline{28.23} & \textbf{35.31} & 36.62 & \textbf{55.62} &
10.61 & 10.40 & \underline{20.71} & \textbf{41.33} & 37.79 & \textbf{48.13} \\
256k &
24.73 & 30.77 &
14.52 & 13.75 & \underline{25.50} & \textbf{38.04} & 43.04 & \textbf{61.39} &
12.79 & 10.86 & \underline{25.52} & \textbf{38.79} & 44.91 & \textbf{57.49} \\
512k &
22.23 & 31.21 &
7.48 & 7.97 & \textbf{32.93} & \underline{31.57} & 35.64 & \textbf{59.59} &
12.94 & 10.12 & \underline{21.49} & \textbf{40.14} & 35.77 & \textbf{55.26} \\
1M   &
17.70 & 30.45 &
8.30 & 3.80 & \underline{25.62} & \textbf{34.20} & 35.91 & \textbf{56.86} &
3.15 & 2.55 & \underline{21.77} & \textbf{41.49} & 38.40 & \textbf{58.57} \\
		
\midrule
\multicolumn{15}{l}{\textbf{2Wiki}} \\
32k  &
37.19 & 37.25 &
49.71 & 39.58 & \textbf{55.62} & \underline{54.70} & 56.43 & \textbf{70.66} &
37.41 & 39.20 & \underline{44.71} & \textbf{44.52} & 49.57 & \textbf{68.78} \\
64k  &
32.52 & 31.54 &
40.86 & 28.34 & \underline{47.56} & \textbf{49.88} & 48.55 & \textbf{74.84} &
42.00 & 37.38 & \underline{40.06} & \textbf{49.53} & 51.68 & \textbf{64.80} \\
128k &
34.96 & 36.91 &
39.72 & 30.47 & \textbf{50.27} & \underline{47.40} & 46.18 & \textbf{70.46} &
33.51 & 29.75 & \underline{46.31} & \textbf{50.60} & 49.66 & \textbf{61.88} \\
256k &
22.86 & 32.26 &
20.39 & 20.02 & \underline{43.99} & \textbf{46.20} & 41.42 & \textbf{66.62} &
21.67 & 13.96 & \underline{34.85} & \textbf{48.15} & 47.73 & \textbf{63.55} \\
512k &
15.12 & 36.73 &
20.16 & 22.52 & \underline{48.23} & \textbf{50.81} & 39.09 & \textbf{71.54} &
22.77 & 15.66 & \underline{32.93} & \textbf{54.74} & 46.31 & \textbf{65.19} \\
1M   &
17.13 & 27.77 &
19.12 & 17.30 & \underline{45.45} & \textbf{45.55} & 35.18 & \textbf{66.39} &
12.15 & 5.60 & \underline{28.49} & \textbf{50.62} & 43.18 & \textbf{68.20} \\

\bottomrule
\end{tabular}}

\label{tab:generality_backbones}
\end{table*}

%% file: tables/longbench_task_T.tex
\newcolumntype{Y}{>{\raggedright\arraybackslash}X}

\newcommand{\barscale}{0.45pt} 
\newcommand{\barheight}{3.5pt}
\newcommand{\barraise}{\dimexpr0.5ex - 0.5\dimexpr\barheight\relax\relax}

\newcommand{\incLite}[1]{%
    \mbox{\color{teal!80}\tiny~+#1%
    \hspace{1pt}%
    {\color{teal!30}\rule[\barraise]{\dimexpr#1\dimexpr\barscale\relax\relax}{\barheight}}}%
}
\newcommand{\decLite}[1]{%
    \mbox{\color{red!80}\tiny~#1%
    \hspace{1pt}%
    {\color{red!20}\rule[\barraise]{\dimexpr-#1\dimexpr\barscale\relax\relax}{\barheight}}}%
}

\newcommand{\incDeep}[1]{%
    \mbox{\color{teal}\tiny~+#1%
    \hspace{1pt}%
    {\color{teal!80}\rule[\barraise]{\dimexpr#1\dimexpr\barscale\relax\relax}{\barheight}}}%
}
\newcommand{\decDeep}[1]{%
    \mbox{\color{red}\tiny~#1%
    \hspace{1pt}%
    {\color{red!80}\rule[\barraise]{\dimexpr-#1\dimexpr\barscale\relax\relax}{\barheight}}}%
}

\begin{table*}[t!]

\centering
\scriptsize
\renewcommand{\arraystretch}{1.25} 
\setlength{\tabcolsep}{1.5pt} 
\caption{\textbf{Performance comparison on the LongBench QA benchmark.} We evaluate Qwen series models across five QA datasets. The colored bars indicate the absolute performance gain (\textcolor{teal}{green}) or loss (\textcolor{red}{red}) compared to the YaRN baseline. 
\textbf{InfMem} and its RL variant (highlighted in gray) consistently outperform other methods across all model scales.}
\resizebox{0.95\textwidth}{!}{
\begin{tabularx}{\textwidth}{ll|YYYYY|Y}
\toprule
\multirow{2}{*}{\textbf{Model}} &
\multirow{2}{*}{\textbf{Method}} &
\multicolumn{5}{c|}{\textbf{LongBench QA}} &
\multirow{2}{*}{\textbf{avg}} \\
\cmidrule(lr){3-7}
& & \textbf{NQA} & \textbf{HQA} & \textbf{2Wiki} & \textbf{Qasper} & \textbf{Musique} & \\
\midrule

\multirow{5}{*}{\textbf{Qwen3-1.7B}}
& YaRN      & \underline{17.09} & 33.87 & \underline{50.32} & \textbf{37.91} & \underline{23.86} & \underline{32.61} \\
& MemAgent  & 15.04\decLite{-2.05} & \underline{41.68}\incLite{7.81} & 34.04\decLite{-16.28} & 30.94\decLite{-6.97} & 19.84\decLite{-4.02} & 28.31\decLite{-4.30} \\

& \cellcolor{gray!15}\textbf{InfMem}
  & \cellcolor{gray!15}\textbf{20.25}\incDeep{3.16}
  & \cellcolor{gray!15}\textbf{48.73}\incDeep{14.86}
  & \cellcolor{gray!15}\textbf{54.05}\incDeep{3.73}
  & \cellcolor{gray!15}\underline{33.91}\decDeep{-4.00}
  & \cellcolor{gray!15}\textbf{28.40}\incDeep{4.54}
  & \cellcolor{gray!15}\textbf{37.07}\incDeep{4.46} \\

\cmidrule(lr){2-8}
& MemAgent +RL
  & 19.23\incLite{2.14} & 50.22\incLite{16.35} & 47.58\decLite{-2.74} & \textbf{35.48}\decLite{-2.43} & 30.35\incLite{6.49} & 35.90\incLite{3.29} \\

& \cellcolor{gray!30}\textbf{InfMem +RL}
  & \cellcolor{gray!30}19.23\incDeep{2.14}
  & \cellcolor{gray!30}\textbf{59.28}\incDeep{25.41}
  & \cellcolor{gray!30}\textbf{55.02}\incDeep{4.70}
  & \cellcolor{gray!30}33.19\decDeep{-4.72}
  & \cellcolor{gray!30}\textbf{40.98}\incDeep{17.12}
  & \cellcolor{gray!30}\textbf{41.54}\incDeep{8.93} \\

\midrule
\multirow{5}{*}{\textbf{Qwen3-4B}}
& YaRN      & \underline{21.46} & 53.20 & 50.31 & \textbf{40.14} & \underline{32.18} & 39.46 \\
& MemAgent  & 20.22\decLite{-1.24} & \underline{57.67}\incLite{4.47} & \underline{59.09}\incLite{8.78} & 33.52\decLite{-6.62} & 32.12\decLite{-0.06} & \underline{40.52}\incLite{1.06} \\

& \cellcolor{gray!15}\textbf{InfMem}
  & \cellcolor{gray!15}\textbf{23.27}\incDeep{1.81}
  & \cellcolor{gray!15}\textbf{60.96}\incDeep{7.76}
  & \cellcolor{gray!15}\textbf{69.66}\incDeep{19.35}
  & \cellcolor{gray!15}\underline{35.14}\decDeep{-5.00}
  & \cellcolor{gray!15}\textbf{44.19}\incDeep{12.01}
  & \cellcolor{gray!15}\textbf{46.64}\incDeep{7.18} \\

\cmidrule(lr){2-8}
& MemAgent +RL
  & 20.74\decLite{-0.72} & 63.80\incLite{10.60} & 67.83\incLite{17.52} & \textbf{41.02}\incLite{0.88} & 42.14\incLite{9.96} & 47.11\incLite{7.65} \\

& \cellcolor{gray!30}\textbf{InfMem +RL}
  & \cellcolor{gray!30}\textbf{20.77}\decDeep{-0.69}
  & \cellcolor{gray!30}\textbf{65.14}\incDeep{11.94}
  & \cellcolor{gray!30}\textbf{74.76}\incDeep{24.45}
  & \cellcolor{gray!30}40.74\incDeep{0.60}
  & \cellcolor{gray!30}\textbf{53.22}\incDeep{21.04}
  & \cellcolor{gray!30}\textbf{50.93}\incDeep{11.47} \\

\midrule
\multirow{5}{*}{\textbf{Qwen2.5-7B}}
& YaRN      & 16.12 & 42.92 & 40.55 & 28.84 & 19.28 & 29.54 \\
& \textbf{MemAgent}  & \textbf{19.86}\incLite{3.74} & \textbf{53.23}\incLite{10.31} & \textbf{55.40}\incLite{14.85} & \textbf{31.63}\incLite{2.79} & \textbf{36.52}\incLite{17.24} & \textbf{39.33}\incLite{9.79} \\

& \cellcolor{gray!15}InfMem
  & \cellcolor{gray!15}\underline{19.76}\incDeep{3.64}
  & \cellcolor{gray!15}\underline{52.95}\incDeep{10.03}
  & \cellcolor{gray!15}\underline{48.78}\incDeep{8.23}
  & \cellcolor{gray!15}\underline{31.09}\incDeep{2.25}
  & \cellcolor{gray!15}\underline{31.69}\incDeep{12.41}
  & \cellcolor{gray!15}\underline{36.85}\incDeep{7.31} \\

\cmidrule(lr){2-8}
& MemAgent +RL
  & 19.47\incLite{3.35} & 56.17\incLite{13.25} & 57.66\incLite{17.11} & 35.52\incLite{6.68} & 31.23\incLite{11.95} & 40.01\incLite{10.47} \\

& \cellcolor{gray!30}\textbf{InfMem +RL}
  & \cellcolor{gray!30}\textbf{20.43}\incDeep{4.31}
  & \cellcolor{gray!30}\textbf{60.34}\incDeep{17.42}
  & \cellcolor{gray!30}\textbf{65.19}\incDeep{24.64}
  & \cellcolor{gray!30}\textbf{35.68}\incDeep{6.84}
  & \cellcolor{gray!30}\textbf{50.66}\incDeep{31.38}
  & \cellcolor{gray!30}\textbf{46.46}\incDeep{16.92} \\

\bottomrule
\end{tabularx}}
\label{tab:longbench_delta_table}
\end{table*}
\vspace{-0.2cm}

%% file: tables/token_cost.tex
\begin{table}[h!]
\centering
\small
\setlength{\tabcolsep}{6pt}
\renewcommand{\arraystretch}{1.12}
\caption{\textbf{Decode (generated) tokens per query} at a short (length 800, ctx $\approx$112k) and a long (length 6400, ctx $\approx$896k) setting.
Among the methods that scale to ultra-long context, InfMem is far cheaper than MemAgent and grows much more slowly ($\approx5\times$ vs.\ $\approx11\times$ from 800 to 6400).}
\label{tab:token_cost}
\begin{tabular}{ll|rr}
\toprule
\textbf{Method} & \textbf{Backbone} & \textbf{@800} & \textbf{@6400} \\
\midrule
Mem-$\alpha$                     & 4B  & 15{,}994 & 114{,}992 \\
MemAgent                         & 4B  & 10{,}697 & 117{,}418 \\
\rowcolor{gray!15}
\textbf{InfMem (ours)}           & 4B  & \textbf{6{,}728} & \textbf{33{,}085} \\\bottomrule
\end{tabular}

\vspace{2pt}
\end{table}

%% file: sections/7_conclusion.tex
\section{Conclusion}
\label{sec:conclusion}
In this work, we present InfMem, a cognitive agent designed to resolve the fidelity dilemma in ultra-long context reasoning. By integrating structured evidence management with a robust SFT$\rightarrow$RL training pipeline, InfMem excels in long-horizon search and retrieval. Empirical evaluations on 1M-token benchmarks demonstrate that InfMem outperforms the state-of-the-art MemAgent with double-digit accuracy improvements across various Qwen models, while simultaneously reducing latency by 3.9$\times$ via inference early stopping. Our findings suggest that as context windows scale, the primary bottleneck shifts from raw memory capacity to cognitive control: the ability to effectively discern and "know what is known".

%% file: sections/8_statement.tex
\section*{Ethics Statement}
This work proposes a method for improving long-document question answering under bounded compute and memory budgets.
The approach does not introduce new datasets, collect personal data, or target high-risk application domains.
Potential benefits include improved efficiency and reliability in document analysis tasks such as technical review and knowledge synthesis. In this work, LLMs were used solely for grammar checking and textual correcting, consistent with accepted usage guidelines.

%% file: sections/X_appendix.tex
\input{appendix/implementation}

\input{appendix/case_study}

\input{appendix/ablation}

\input{appendix/extended_results}

%% file: appendix/implementation.tex
\section{Implementation}
\subsection{Prompts and Templates}
\label{app:prompt}

We use two structured templates to implement the recurrent \textsc{Retrieve--Compress} loop:
a \textbf{Retriever Template} for decision making and query formation, and a \textbf{Memory Template} for faithful evidence compression.

\paragraph{Retriever Template (Figure~\ref{fig:retrieval_planner_template})}
The retriever prompt conditions on the current question and the accumulated memory, and asks the model to (i) assess whether the memory already contains sufficient evidence to answer, and (ii) if not, produce a function-call specification for external retrieval.
Concretely, the template outputs a discrete decision (\textsc{STOP} vs.\ \textsc{RETRIEVE}); when retrieval is needed, it emits a search \texttt{query} and a \texttt{top\_k} value.
This design turns retrieval into an explicit, controllable action: the model is encouraged to issue broad queries when evidence is missing, refine queries when retrieval results are noisy or mismatched, and allocate \texttt{top\_k} based on uncertainty (larger $k$ when multiple candidate entities/facts exist; smaller $k$ when the target is specific).
By tying retrieval decisions to the evolving memory state, the agent can avoid redundant searches and terminate early once decisive evidence has been accumulated.

\paragraph{Memory Template (Figure~\ref{fig:recurrent_memory_template})}
The memory prompt performs bounded, evidence-centric compression .
At each step, it is given two sources: (1) the newly retrieved chunk (high-relevance but potentially noisy) and (2) a recurrent chunk from the running context (stable but may be redundant).
The template instructs the model to extract only answer-relevant facts, normalize entities/aliases, and write a compact memory update that preserves verifiable evidence (names, dates, titles, and relations) while discarding stylistic or speculative content.
Importantly, the template enforces \emph{selective} compression across the two inputs: it prioritizes new complementary evidence from retrieval, but retains previously stored facts when they remain useful, preventing memory drift and uncontrolled growth.

\subsection{Data Construction Details}
\label{app:data}
\paragraph{Unified long-context synthesis pipeline.}
All synthesized long-context QA instances share the same supervision format: a question $Q$ and an answer $A$, together with a set of \emph{gold evidence documents} (or paragraphs) annotated by the source dataset.
We convert each original instance into a \emph{single long document} by mixing (i) the gold evidence documents, and (ii) a large pool of \emph{distractor} documents sampled from the same corpus.
Concretely, for each instance we build three text pools: the query ($Q$), the evidence set ($\mathcal{D}_{\text{gold}}$), and a distractor pool ($\mathcal{D}_{\text{dist}}$) drawn from the dataset's training corpus.\footnote{We sample distractors from the same corpus to preserve domain/style match, making the task harder than using out-of-domain noise.}
We then create a candidate document list by shuffling documents with a fixed random seed, insert each gold document \emph{exactly once} at the document level, and keep appending distractors until reaching a target token budget.
This yields a \emph{controlled} setting where (1) the answer is always supported by $\mathcal{D}_{\text{gold}}$, while (2) retrieval difficulty scales with the number of distractors and total context length.

\subsubsection{Cold-start SFT data}
\label{app:sft_data}
Following the NIAH-style long-context QA construction in MemAgent, we synthesize cold-start SFT data from three QA sources: HotpotQA, SQuAD, and MuSiQue.
Each source contributes 4,096 instances sampled from its training split.
For each instance, we construct a long document at a fixed target length (32K tokens) by iteratively inserting distractor documents until the budget is met.\footnote{In practice, we first \emph{pre-scan} candidate distractor documents to determine how many whole documents can be inserted under a given token budget.
We then construct the long document in a single pass by inserting all gold evidence documents once and appending the maximal number of distractors without exceeding the target length, truncating only at document boundaries.
}
We use \texttt{Qwen3-32B} as the teacher with \emph{thinking enabled} to generate protocol-consistent interaction traces under our \textsc{PreThink--Retrieve--Write} workflow:
the teacher (i) plans and emits structured retrieve calls, (ii) updates a bounded agent memory by writing compressed evidence, and (iii) decides when to stop retrieving and answer.
We then distill student backbones (Qwen3-1.7B, Qwen3-4B, and Qwen2.5-7B-Instruct) on these trajectories.

\paragraph{Question decompositions.}
MuSiQue provides an optional question decomposition (multi-hop sub-questions).
We feed decompositions \emph{only to the teacher} to elicit cleaner and more stable planning traces; students never observe decompositions, gold document IDs, or any teacher-side annotations during either training or inference.
For HotpotQA and SQuAD, the teacher autonomously decides whether to decompose the question in its private reasoning and how to formulate retrieval queries.

\paragraph{Trajectory filtering.}
To ensure supervision quality, we retain only traces whose final answers are correct under the official evaluation protocol of the underlying dataset and discard all failed attempts. We additionally remove excessively long traces that would exceed the memory budget or truncate the agent memory/state; this ensures the student is trained on trajectories that are feasible at inference time under the same bounded-memory constraints.

After this filtering process, we decompose the successful trajectories into individual turns, resulting in a total of 29,717 single-turn dialogue instances. These instances constitute our final SFT dataset for training the student backbones.
\subsubsection{RL training data.}
\label{app:RL_data}
For RL training, we utilize the same synthesis pipeline to extend the context length of HotpotQA instances to approximately 28K tokens. We retain the original question-answer pairs while scaling the retrieval difficulty through the insertion of distractors. During the reinforcement learning phase, the model is optimized using the Exact Match (EM) score between the generated response and the ground-truth answer as the primary reward signal. This setup ensures that the environment remains consistent with our SFT stage, allowing the RL process to focus specifically on refining the agent’s decision-making—such as retrieval timing and memory management—under long-context constraints.
\input{figs/retriever_template}

\input{figs/memory_template}

\input{figs/sft_prethink_matter}

\input{figs/bef_aft_RL}

\input{figs/learn_topk}

\subsubsection{Evaluation Benchmark}
\label{app:bench}
\paragraph{Synthesized long-context QA benchmarks (extreme scaling).}
To evaluate robustness under extreme context scaling, we create long-document variants following the NIAH-style construction for representative multi-hop QA tasks, including HotpotQA, 2WikiMultihopQA, and MuSiQue; we also include the synthetic SQuAD setting used in MemAgent for direct comparison.
We use each dataset's \emph{test split} and sample 128 instances per task.
For each fixed question set, we generate multiple test variants at increasing target lengths (e.g., 32K/28K, 64K/56K, 128K/112K, up to 1M/896K tokens) by progressively inserting more distractors while keeping the gold evidence set unchanged.
Gold evidence is inserted once per instance at the document level with a fixed seed, and distractors are sampled from the same corpus to preserve distributional match.
This protocol ensures that differences across lengths reflect only the effect of \emph{context scaling} (more distractors / longer inputs), not changes in questions or evidence.

\paragraph{Task-specific token budgets.}
The minimum target length differs slightly across tasks: HotpotQA uses 28K tokens to match the document-count-based construction inherited from the RL dataset, while other tasks use fixed token budgets (32K/64K/128K/$\dots$/1M) and insert as many whole documents as allowed under each budget.

\paragraph{LongBench QA benchmarks (natural distributions).}
To verify transfer beyond synthetic distractor insertions, we additionally evaluate on LongBench QA using its original documents and distributions.
We report F1 on NarrativeQA, HotpotQA, 2WikiMultihopQA, Qasper, and MuSiQue following the official LongBench evaluation protocol.

\subsection{Training setup}
\label{app:train_set}

\subsubsection{Model Configuration and Baselines}
\label{app:model_config}
To ensure a rigorous evaluation, we standardize the recurrent interaction settings across both \textsc{InfMem} and the baseline \textsc{MemAgent} (based on Qwen-1.5B/4B backbones).
\paragraph{Recurrent Processing Setup.}Both models operate with a fixed \textbf{recurrent chunk size of 5,000 tokens}. To maintain consistency in the reasoning horizon, we align the \textbf{maximum generation length} (1.5k tokens) and the interaction iteration steps for both models. For \textsc{InfMem}, we enable BM25-based retrieval with a cap of 4,000 retrieved tokens per step. Crucially, during the memory update phase of \textsc{InfMem}, we explicitly filter out reasoning/thinking steps, retaining only the schema-consistent memory tokens to maximize information density.
\paragraph{Baseline Fairness.}For the \textsc{MemAgent} reproduction, we disable the optional ``thinking mode'' (as discussed in \S\ref{app:ablation_baseline}) to adhere to its standard efficient setting.It is important to note that our comparison aligns the \emph{output} constraints (generation length and steps) rather than the input/memory budget. Since \textsc{InfMem} processes additional retrieved context (up to 4k tokens) within the same iteration framework, it is required to compress a significantly larger volume of information into the memory state compared to MemAgent.This setup ensures we are not weakening the baseline; rather, we are testing \textsc{InfMem}'s ability to handle higher information loads under strictly bounded generation resources.

\section{Baseline Implementation Details}
\label{app:baseline_details}

To ensure a fair and rigorous comparison against our proposed methods, we carefully reconstructed the experimental setups for the strong agentic baselines, RESP and Mem-$\alpha$. Because these methods were originally designed for different scales or exhibited generation bottlenecks, we integrated a larger Qwen3-32B backbone to maximize their competitive performance. We also standardized the configurations for our standard long-context and retrieval baselines.

\textbf{RESP:} We strictly follow the implementation details provided in the original paper. To accurately replicate their iterative RAG logic, we utilized their exact prompt templates. We employed Qwen3-32B as the backbone model throughout the entire process, handling the iterative reasoning, planning, summarization, and the final generation.

\textbf{Mem-$\alpha$:} We integrated the publicly available Mem-$\alpha$ implementation directly into our evaluation framework. Specifically, we utilized their released 4B checkpoint to serve as the context manager. However, empirical observations indicated that the RL-tuned Mem-$\alpha$ checkpoint exhibited suboptimal generative question-answering capabilities. To prevent the baseline's performance from being unfairly bottlenecked by poor generation rather than poor context management, we decoupled the architecture: we delegated the final QA generation phase to Qwen3-32B while retaining the 4B model strictly for context management.

\textbf{YaRN:} For evaluations utilizing YaRN, we apply the official RoPE scaling configurations established for the Qwen2.5 and Qwen3 models. Specifically, we set the \texttt{rope\_type} to \texttt{yarn}, apply a scaling \texttt{factor} of 4.0, and maintain the \texttt{original\_max\_position\_embeddings} at 32,768. 

\textbf{RAG:} For the standard RAG baseline, our reported main results utilize a chunk size of 1,000 tokens and retrieve the top-6 chunks. To determine this optimal configuration, we conducted an ablation study on the synthesized HQA benchmark across context lengths of 28k, 56k, and 112k using Qwen3-4B. We evaluated multiple retrieval variants, including top-12 with 1,000-token chunks, top-6 with 2,000-token chunks, top-2 with 1,000-token chunks, and top-6 with 1,000-token chunks. Empirical results demonstrated that retrieving the top-6 chunks with a 1,000-token chunk size consistently yielded slightly better performance, and it was thus selected as the default configuration for all RAG evaluations.
\begin{table}[t]
\centering
\small
\setlength{\tabcolsep}{6pt}
\renewcommand{\arraystretch}{1.2}

\caption{Additional details of InfMem inference protocol.}
\label{tab:infmem_appendix}

\begin{subtable}{\linewidth}
\centering
\caption{InfMem inference algorithm}
\label{tab:infmem_alg}

\begin{tabular}{p{0.96\linewidth}}
\toprule
\textbf{Algorithm 1: InfMem Inference Protocol} \\ \midrule
\textbf{Input:} question $q$; streaming chunks $\{c_t\}_{t=1}^T$; global retrieval units $\{p_j\}_{j=1}^N$; budget $M$ \\
\textbf{Initialize:} memory $m_0 \leftarrow \emptyset$ \\[2pt]
\textbf{for} $t = 1$ \textbf{to} $T$ \textbf{do} \\
    \quad \color{gray}// Step 1: Monitor \& Plan (PreThink) \\
    \quad $(a_t, u_t, k_t) \leftarrow \textsc{PreThink}(q, m_{t-1})$ \\
    
    \quad \textbf{if} $a_t = \textsc{STOP}$ \textbf{then} \\
    \quad \quad \textbf{break} \quad \color{gray}// Early stopping triggered \\
    \quad \textbf{end if} \\[2pt]
    
    \quad \color{gray}// Step 2: Seek (Retrieve) \\
    \quad \textbf{if} $a_t = \textsc{RETRIEVE}$ \textbf{then} \\
    \quad \quad $r_t \leftarrow \textsc{Retrieve}(u_t, k_t; \{p_j\})$ \\
    \quad \textbf{end if} \\[2pt]
    
    \quad \color{gray}// Step 3: Update (Write with Joint Compression) \\
    \quad $m_t \leftarrow \textsc{Write}(q, m_{t-1}, c_t, r_t; M)$ \\
\textbf{end for} \\[4pt]

\color{gray}// Final Answer Generation \\
$\hat{y} \leftarrow \textsc{Answer}(q, m_{\text{final}})$ \\ \bottomrule
\end{tabular}
\end{subtable}

\vspace{8pt}

\begin{subtable}{\linewidth}
\centering
\caption{Design rationale of InfMem components}
\label{tab:infmem_rationale}

\begin{tabular}{p{0.28\linewidth} p{0.66\linewidth}}
\toprule
\textbf{Component} & \textbf{Rationale} \\ \midrule
\textsc{PreThink} & Acts as a state-dependent controller to monitor sufficiency and plan query $u_t$ based on memory $m_{t-1}$ \\
\textsc{Retrieve} & Enables global, non-monotonic access to sparse evidence $\{p_j\}$ missed by linear scanning \\
\textsc{Write} & Performs evidence-aware \emph{joint compression}, prioritizing bridging links from both $c_t$ and $r_t$ \\
\textsc{Early Stop} & Terminates inference once evidence is sufficient ($a_t=\textsc{STOP}$), reducing latency and redundancy \\
\bottomrule
\end{tabular}
\end{subtable}

\end{table}
\input{tables/sft_RL_detail}
\input{tables/train_free_sft_RL}

%% file: figs/retriever_template.tex
\begin{figure*}[t]
\centering
\scriptsize

\begin{bigcaseblue}{Retrieval Planner Prompt Template}
\cssec{Template}

\texttt{You are a Retrieval Planner.}\par
\par

\texttt{Your ONLY task is to decide whether to perform another retrieval using `retrievesearch`, or STOP retrieving.}\par
\par

\texttt{You MUST NOT answer the QUESTION.}\par
\texttt{Another model will use MEMORY to answer later.}\par
\par

\texttt{Guidelines:}\par
\texttt{- Retrieval is cheap. Unless MEMORY clearly contains all essential information, you are encouraged to retrieve.}\par
\texttt{- You may retrieve multiple times. At each step, refine your search direction.}\par
\texttt{- Avoid repeating any previous queries in RETRIEVAL\_HISTORY (unless meaningfully refined).}\par
\par

\texttt{When deciding whether to retrieve again:}\par
\texttt{1. Break the QUESTION into specific sub-questions or information needs.}\par
\texttt{2. Compare these needs with what MEMORY already contains.}\par
\texttt{3. Identify which facts are still missing, uncertain, or incomplete.}\par
\texttt{4. If something important is missing, design a NEW search query focused only on that missing information.}\par
\texttt{   - You may explore related clues hinted in MEMORY.}\par
\texttt{   - Queries should be concise, specific, and actionable.}\par
\texttt{5. If MEMORY already contains all necessary information, choose to STOP.}\par
\par

\texttt{If you choose retrieval, you MUST output a function call to `retrievesearch` with:}\par
\texttt{- a new `query` (different from RETRIEVAL\_HISTORY unless refined),}\par
\texttt{- and a `top\_k` suited to your confidence (small: focused; large: broad exploration).}\par
\texttt{- In early retrieval steps, you may exlore more documents.}\par
\texttt{- In later steps, focus on refining MEMORY.}\par
\par

\texttt{ORIGINAL QUESTION:}\par
\texttt{\{prompt\}}\par
\par

\texttt{\textless retrieval\_history\textgreater}\par
\texttt{\{retrieval\_history\}}\par
\texttt{\textless /retrieval\_history\textgreater}\par
\par

\texttt{CURRENT MEMORY:}\par
\texttt{\{memory\}}\par
\par

\end{bigcaseblue}

\caption{Prompt template for the Retrieval Planner, which decides whether to call \texttt{retrievesearch} again or stop, without answering the question.}
\label{fig:retrieval_planner_template}
\end{figure*}

%% file: figs/memory_template.tex
\begin{figure*}[t]
\centering
\scriptsize

\begin{bigcaseblue}{Memory Update Prompt Template}
\cssec{Template}

\texttt{You are presented with a problem, a section of an article that may contain the answer to the problem, and a previous memory. Please read the provided section carefully and update the memory with the new information that helps to answer the problem.}\par
\par

\texttt{The given section has two parts. One is a retrieved chunk, which is retrieved by the given question. Another is the recurrent chunk which is provided recurrently. Both chunks might contain useful information, while the retrieved chunk may have a higher chance.}\par
\par

\texttt{\textless problem\textgreater}\par
\texttt{\{prompt\}}\par
\texttt{\textless /problem\textgreater}\par
\par

\texttt{\textless retrieved\_chunk\textgreater}\par
\texttt{\{retrieve\}}\par
\texttt{\textless /retrieved\_chunk\textgreater}\par
\par

\texttt{\textless recurrent\_chunk\textgreater}\par
\texttt{\{chunk\}}\par
\texttt{\textless /recurrent\_chunk\textgreater}\par
\par

\texttt{\textless memory\textgreater}\par
\texttt{\{memory\}}\par
\texttt{\textless /memory\textgreater}\par
\par

\texttt{Updated memory:}\par

\end{bigcaseblue}

\caption{Prompt template for memory updating, integrating both retrieved and recurrent chunks to refine the memory state.}
\label{fig:recurrent_memory_template}
\end{figure*}

%% file: figs/sft_prethink_matter.tex
\begin{figure*}[t]
\centering
\setlength{\tabcolsep}{6pt} 
\scriptsize

\begin{tcbraster}[
  raster columns=2,
  raster column skip=8pt,
  raster equal height,
  raster valign=stretch
]

\begin{bigcase}{No-think (question-to-query copying)}
\cssec{Original question \& correct answer}
\textbf{Q:} Which state does Sang-Wook Cheong work in as a materials scientist? \par
\textbf{A:} New Jersey. \par

\cssec{Previous retrieval history}
Step 1: query='Sang-Wook Cheong materials scientist state', top\_k=5 \par
Step 2: query='Sang-Wook Cheong materials scientist state', top\_k=5 \par
Step 3: query='Sang-Wook Cheong materials scientist state', top\_k=5 \par

\cssec{Function call}
FUNCTION: retrievesearch \par
ARGS: \{"query": "Sang-Wook Cheong materials scientist state", "top\_k": 5\} \par

\cssec{Memory update}
Updated memory: \par
Sang-Wook Cheong is a Korean American materials scientist at Rutgers University. \par
The state where he works is New Jersey. \par
\end{bigcase}%
\begin{bigcaseblue}{Think (SFT \textsc{PreThink}: plan with history \& memory)}
\cssec{Original question \& correct answer}
\textbf{Q:} Where is the company headquarters that makes the GU230 mobile phone? \par
\textbf{A:} Yeouido-dong, Seoul, South Korea. \par

\cssec{Previous retrieval history}
Step 1: query='GU230 mobile phone manufacturer', top\_k=5 \par
Step 2: query='GU230 mobile phone company headquarters', top\_k=5 \par

\cssec{Function call (planned follow-up)}
\textless think\textgreater \par
 Okay, let's see. The user is asking where the headquarters of the company that makes the GU230 mobile phone is located. The retrieval history shows that two searches have been done so far.\par
 ...\par
 next query should focus on confirming the relationship between M.Mobile and the GU230 model, maybe using "M.Mobile GU230 manufacturer" to see if there's a direct link in the documents. \par
 \textless /think\textgreater

FUNCTION: retrievesearch \par
ARGS: \{"query": "M.Mobile subsidiaries and mobile phone production", "top\_k": 5\} \par

\cssec{Memory update (evidence composition)}
\textless think\textgreater  \par
 Okay, let's tackle this problem step by step. The user is asking where \par
 ... \par
 does mention LG as the manufacturer. The prior analysis missed that. So the correct answer is the headquarters of LG Electronics as per Document 96. \par

 \textless /think\textgreater

Updated memory: \par
GU230 is manufactured by LG Electronics. \par
LG Electronics is headquartered in Yeouido-dong, Seoul, South Korea. \par
\end{bigcaseblue}
\end{tcbraster}

\caption{\textbf{Visualized retrieval trajectories: No-think vs Think.}
Without \textsc{PreThink}, the model tends to copy the question into repetitive queries.
After SFT, the planner conditions on retrieval history and memory, identifies missing links, and issues a targeted follow-up query to complete multi-hop evidence composition.}
\label{fig:think_vs_nothink}
\end{figure*}

%% file: figs/bef_aft_RL.tex
\begin{figure*}[t]
\centering
\scriptsize

\begin{tcbraster}[
  raster columns=2,
  raster column skip=8pt,
  raster equal height,
  raster valign=stretch
]

\begin{bigcase}{Before RL (retrieval strategy preserved, but fails to extract key evidence)}
\cssec{Original question \& correct answer}
\textbf{Q:} which former lawyer previously held the seat held by Yuh-Line Niou \par
\textbf{A:} Sheldon Silver \par

\cssec{Previous retrieval history}
Step 1: query='Yuh-Line Niou seat current occupant', top\_k=5 \par
Step 2: query='Yuh-Line Niou seat previous occupants', top\_k=5 \par
Step 3: query='Frank Lautenberg career', top\_k=5 \par

\cssec{Function call}
FUNCTION: retrievesearch \par
ARGS: \{"query": "Yuh-Line Niou seat previous occupants", "top\_k": 5\} \par

\cssec{Memory update (incorrect composition)}
Updated memory: \par
The seat previously held by Yuh-Line Niou is linked to New Jersey's Senate history. \par
Frank Lautenberg succeeded a retiring Republican senator in the 1982 New Jersey Senate election. \par
Thus, Frank Lautenberg is the most plausible former lawyer for the seat. \par

\cssec{Answer (wrong)}
Frank Lautenberg \par
\end{bigcase}%
\begin{bigcaseblue}{After RL (same retrieve pattern, but extracts the key sentence and composes correctly)}
\cssec{Original question \& correct answer}
\textbf{Q:} which former lawyer previously held the seat held by Yuh-Line Niou \par
\textbf{A:} Sheldon Silver \par

\cssec{Previous retrieval history}
Step 1: query='Yuh-Line Niou seat', top\_k=5 \par
Step 2: query='seat held by Yuh-Line Niou previous holder', top\_k=5 \par
Step 3: query='Yuh-Line Niou seat holder', top\_k=5 \par

\cssec{Function call}
FUNCTION: retrievesearch \par
ARGS: \{"query": "seat held by Yuh-Line Niou previous holder", "top\_k": 5\} \par

\cssec{Memory update (key evidence extracted)}
Updated memory: \par
Yuh-Line Niou serves in the New York State Assembly for the 65th district. \par
The 65th district seat was previously held by \textbf{Sheldon Silver}. \par
Sheldon Silver is a \textbf{former lawyer}. \par

\cssec{Answer (correct)}
Sheldon Silver \par
\end{bigcaseblue}

\end{tcbraster}

\caption{\textbf{RL effect on evidence extraction and memory writing.}
Both runs use a similar retrieval pattern, but \emph{before RL} the agent fails to identify the direct sentence linking Niou's seat to the previous holder and instead hallucinates an unrelated political chain.
\emph{After RL}, the agent reliably extracts the decisive evidence (Niou $\rightarrow$ NY Assembly 65th district $\rightarrow$ Sheldon Silver) and writes a compact, answer-ready memory.}
\label{fig:rl_before_after_case}
\end{figure*}

%% file: figs/learn_topk.tex
\begin{figure*}[t]
\centering
\scriptsize

\begin{tcbraster}[
  raster columns=2,
  raster column skip=8pt,
  raster equal height,
  raster valign=stretch
]

\begin{bigcase}{Before RL (uncontrolled exploration, failure to identify the target franchise)}
\cssec{Original question \& correct answer}
\textbf{Q:} How many original games are there in the tactical role-playing video game franchise developed by the same Japanese company who developed the ``Wars'' series of games? \par
\textbf{A:} Twelve \par

\cssec{Previous retrieval history}
Step 3: query=`Intelligent Systems other RPG franchises', top\_k=5 \par
Step 4: query=`Intelligent Systems other franchises', top\_k=3 \par

\cssec{Function call}
\textless think\textgreater  \par
Okay, let's break this down. The user is asking how many original games are in the tactical role-playing franchise \par
...\par
 But wait, maybe the memory is missing some information? For example, maybe there's a third game.
  Let's search for other Japanese tactical RPG franchises besides Sakura Wars to validate.\par

\textless /think\textgreater  \par
FUNCTION: retrievesearch \par
ARGS: \{"query": "Japanese tactical RPG franchises besides Sakuno Wars", "top\_k": 4\} \par

\cssec{Memory update (misidentification)}
Updated memory: \par
The ``Wars'' series is developed by Intelligent Systems. \par
No other tactical RPG franchise by Intelligent Systems is explicitly identified. \par
Advance Wars is part of the same series and does not constitute a separate franchise. \par

\cssec{Answer (wrong)}
The franchise has \textbf{1 original game}. \par
\end{bigcase}%
\begin{bigcaseblue}{After RL (top-$k$–aware exploration and correct evidence composition)}
\cssec{Original question \& correct answer}
\textbf{Q:} How many original games are there in the tactical role-playing video game franchise developed by the same Japanese company who developed the ``Wars'' series of games? \par
\textbf{A:} Twelve \par

\cssec{Previous retrieval history}
Step 3: query=`Fire Emblem original game count number', top\_k=5 \par
Step 4: query=`Fire Emblem series game count', top\_k=5 \par

\cssec{Function call}
\textless think\textgreater  \par
Okay, let's see. The user is asking about the number of original games in the tactical role-playing video game franchise \par
... \par
The next step would be to search for the exact count, possibly using a query like "Sakura Wars original game count number" again, but maybe with a different top\_k. Or maybe check if the franchise's official website or a game database has the count. Let's try another search with a slightly different query to see if any of the retrieved chunks mention the number.\par

\textless /think\textgreater  \par
FUNCTION: retrievesearch \par
ARGS: \{"query": "Sakura Wars series game count", "top\_k": 10\} \par

\cssec{Memory update (key facts extracted)}
Updated memory: \par
The ``Wars'' series is developed by \textbf{Intelligent Systems}. \par
\textbf{Fire Emblem} is a tactical role-playing franchise developed by Intelligent Systems. \par
The Fire Emblem series consists of \textbf{twelve original games}, along with remakes and spin-offs. \par

\cssec{Answer (correct)}
The franchise is \textbf{Fire Emblem}, which has \textbf{twelve original games}. \par
\end{bigcaseblue}

\end{tcbraster}

\caption{\textbf{Effect of RL on retrieval control and multi-hop reasoning.}
Before RL, the agent fails to regulate exploration and prematurely concludes that no other tactical RPG franchise exists beyond the ``Wars'' series.
After RL, the agent learns to adapt the retrieval scope via top-$k$ control, successfully identifies \emph{Fire Emblem} as the relevant franchise, and composes the correct numerical answer from explicit evidence.}
\label{fig:rl_topk_fire_emblem}
\end{figure*}

%% file: tables/sft_RL_detail.tex
\begin{table*}[t]
\centering
\scriptsize
\setlength{\tabcolsep}{4.2pt}
\renewcommand{\arraystretch}{1.12}
\caption{\textbf{InfMem results on LongBench QA (LB) and RULER-QA.}
We report per-task LB scores (NQA, HQA, 2Wiki, Qasper, MuSiQue), along with \textbf{avg\_LB} and \textbf{avg\_RULER-QA}.}
\begin{tabular}{ll| 
S[table-format=2.2] S[table-format=2.2] S[table-format=2.2] S[table-format=2.2] S[table-format=2.2]| 
S[table-format=2.3] |S[table-format=2.9]}
\toprule
\textbf{Setting} & \textbf{Model} &
\textbf{LB NQA} & \textbf{LB HQA} & \textbf{LB 2Wiki} & \textbf{LB Qasper} & \textbf{LB MuSiQue} &
\textbf{avg\_LB} & \textbf{avg\_RULER-QA} \\
\midrule
\multicolumn{9}{l}{\textbf{Train-free}} \\
\midrule
& Qwen3-1.7B & 20.25 & 48.73 & 54.05 & 33.91 & 28.40 & 37.068 & 37.70708333 \\
& Qwen2.5-7B & 19.76 & 52.95 & 48.78 & 31.09 & 31.69 & 36.854 & 47.95833333 \\
& Qwen3-4B   & 23.27 & 60.96 & 69.66 & 35.14 & 44.19 & 46.644 & 50.25041667 \\
\midrule
\multicolumn{9}{l}{\textbf{SFT}} \\
\midrule
& Qwen3-1.7B & 18.12 & 47.88 & 46.97 & 31.90 & 31.25 & 35.224 & 43.71708333 \\
& Qwen2.5-7B & 19.95 & 56.46 & 63.23 & 35.31 & 40.07 & 43.004 & 49.30250000 \\
& Qwen3-4B   & 18.71 & 62.19 & 72.13 & 36.09 & 44.90 & 46.804 & 54.85583333 \\
\midrule
\multicolumn{9}{l}{\textbf{RL}} \\
\midrule
& Qwen3-1.7B & 19.23 & 59.28 & 55.02 & 33.19 & 40.98 & 41.540 & 50.84125000 \\
& Qwen2.5-7B & 20.43 & 60.34 & 65.19 & 35.68 & 50.66 & 46.460 & 59.53380952 \\
& Qwen3-4B   & 20.77 & 65.14 & 74.76 & 40.74 & 53.22 & 50.926 & 66.40375000 \\
\bottomrule
\end{tabular}

\label{tab:infmem_lb_ruler_sft_RL}
\end{table*}

%% file: tables/train_free_sft_RL.tex
\begin{table}[t]
\centering
\scriptsize 
\setlength{\tabcolsep}{2pt} 
\renewcommand{\arraystretch}{1.10}

\caption{\textbf{Performance gains across training stages across model scales.}
Values in parentheses indicate absolute improvements ($\Delta$) over the preceding step. Base Model is the starting point.}

\resizebox{\textwidth}{!}{%
\begin{tabular}{
l |
S[table-format=2.2]
S[table-format=2.2] |
S[table-format=2.2]@{\hspace{1pt}}>{\tiny}l
S[table-format=2.2]@{\hspace{1pt}}>{\tiny}l |
S[table-format=2.2]@{\hspace{1pt}}>{\tiny}l
S[table-format=2.2]@{\hspace{1pt}}>{\tiny}l |
S[table-format=2.2]@{\hspace{1pt}}>{\tiny}l
S[table-format=2.2]@{\hspace{1pt}}>{\tiny}l |
S[table-format=2.2]@{\hspace{1pt}}>{\tiny}l
S[table-format=2.2]@{\hspace{1pt}}>{\tiny}l 
}
\toprule
& \multicolumn{2}{c|}{\textbf{Base Model}}
& \multicolumn{4}{c|}{\textbf{Yarn}}
& \multicolumn{4}{c|}{\textbf{Train-free}}
& \multicolumn{4}{c|}{\textbf{SFT}}
& \multicolumn{4}{c}{\textbf{RL}} \\
\cmidrule(lr){2-3}\cmidrule(lr){4-7}\cmidrule(lr){8-11}\cmidrule(lr){12-15}\cmidrule(lr){16-19}
\textbf{Model}
& {avg\_LB} & {avg\_RULER}
& \multicolumn{2}{c}{avg\_LB} & \multicolumn{2}{c|}{avg\_RULER}
& \multicolumn{2}{c}{avg\_LB} & \multicolumn{2}{c|}{avg\_RULER}
& \multicolumn{2}{c}{avg\_LB} & \multicolumn{2}{c|}{avg\_RULER}
& \multicolumn{2}{c}{avg\_LB} & \multicolumn{2}{c}{avg\_RULER} \\
\midrule
Qwen3-1.7B & 25.64 & 2.64 & 32.61 & (\textbf{+6.97}) & 13.38 & (\textbf{+10.74}) & 37.06 & (\textbf{+4.45}) & 37.70 & (\textbf{+24.32}) & 35.22 & (-1.84) & 43.71 & (\textbf{+6.01}) & 41.54 & (\textbf{+6.32}) & 50.84 & (\textbf{+7.13}) \\
Qwen2.5-7B & 32.31 & 0.00 & 29.54 & (-2.77) & 21.41 & (\textbf{+21.41}) & 36.85 & (\textbf{+7.31}) & 47.95 & (\textbf{+26.54}) & 43.00 & (\textbf{+6.15}) & 49.30 & (\textbf{+1.35}) & 46.46 & (\textbf{+3.46}) & 59.53 & (\textbf{+10.23}) \\
Qwen3-4B & 38.39 & 4.30 & 39.46 & (\textbf{+1.07}) & 25.45 & (\textbf{+21.15}) & 46.64 & (\textbf{+7.18}) & 50.20 & (\textbf{+24.75}) & 46.80 & (\textbf{+0.16}) & 54.85 & (\textbf{+4.65}) & 50.92 & (\textbf{+4.12}) & 66.40 & (\textbf{+11.55}) \\
\bottomrule
\end{tabular}%
}
\label{tab:lb_ruler_gains_transposed}
\end{table}

%% file: appendix/case_study.tex
\section{Case Study}
\label{app:case}

\subsection{sft enhance dieversity}

\paragraph{Why SFT warmup is necessary.}
SFT is critical for making tool-use \emph{reliable} in our agentic retrieval loop.
In practice, base backbones do not consistently exhibit disciplined query planning: the smaller Qwen3-1.7B has limited intrinsic reasoning capacity, while the instruction-tuned Qwen2.5-7B still fails to reliably trigger deliberate multi-step planning under our \textsc{Retrieve--Compress} protocol.
Empirically, Table~\ref{tab:infmem_lb_ruler_sft_RL} and Table~\ref{tab:lb_ruler_gains_transposed}  shows consistent improvements from Train-free to SFT across both LB and RULER-QA, and even the strongest backbone (Qwen3-4B) benefits substantially, suggesting that supervised warmup improves not only downstream QA accuracy but also the quality of intermediate actions.\\

Qualitatively, Figure~\ref{fig:think_vs_nothink} visualizes retrieval trajectories on the same instance: without \textsc{PRETHINK}, the model often degenerates into copying the question (or a lightly rewritten variant) as the search query, leading to repetitive, low-information retrievals.
After SFT, the planner conditions on retrieval history and the current memory state, identifies missing links needed for multi-hop composition, and issues targeted follow-up queries, yielding more informative function calls and more dependable evidence aggregation.
\input{figs/early_stop}

\input{tables/InfMem_chunk_size_ablation}

\subsection{RL boost the performance}

RL further strengthens InfMem beyond SFT by explicitly optimizing \emph{long-horizon} tool-use under verifiable QA rewards, yielding gains along two complementary axes:
\textbf{(i) memory compression / evidence writing}, and \textbf{(ii) planning / retrieval control}.\\

As shown in Figure~\ref{fig:rl_before_after_case}, the pre-RL agent may follow a superficially similar retrieval pattern, yet fails at the decisive step of \emph{extracting and committing} the key sentence into memory: rather than grounding “Niou’s seat” in the exact district that links to the previous holder, it writes an unrelated political chain and hallucinates an incorrect former lawyer.
After RL, the agent consistently identifies the decisive evidence chain (Niou $\rightarrow$ NY Assembly 65th district $\rightarrow$ Sheldon Silver) and writes a compact, answer-ready memory, enabling a correct final answer.\\

Figure~\ref{fig:rl_topk_fire_emblem} highlights a complementary improvement in \emph{planning}: RL teaches the agent to regulate exploration by adapting retrieval scope (e.g., via top-$k$ control) instead of drifting or stopping prematurely.
Before RL, the agent prematurely concludes that no tactical RPG franchise exists beyond the “Wars” series; after RL, it expands search when uncertain, discovers \emph{Fire Emblem}, and composes the correct numerical answer from explicit evidence.\\

Taken together, these case studies suggest that RL does not merely increase tool usage; it trains the agent to \emph{write the right information} into memory and to \emph{plan the right next action}—balancing targeted exploration with timely stopping.

\subsection{Early stop}
 Beyond accuracy, early stopping substantially improves inference efficiency. As shown in Fig.~\ref{fig:case_early_stop}, once \textsc{PreThink} determines that the required evidence is already present in memory, the agent explicitly terminates the recurrent retrieve–write loop. This allows the model to exit inference as soon as it is confident in the answer, rather than continuing unnecessary iterations over the remaining context. As a result, inference time is no longer proportional to the document length or number of chunks (i.e., avoiding the typical $O(n)$ recurrent generation cost), and instead approaches constant-time behavior in practice when decisive evidence is found early.

%% file: figs/early_stop.tex
\begin{figure*}[t]
\centering
\scriptsize

\begin{bigcaseblue}{Early Stop (with \textsc{PreThink}: decide retrieval vs.\ stop)}
\cssec{Original question \& correct answer}
\textbf{Q:} Which American composer provided music for the musical ‘Something More!’? \par
\textbf{A:} Sammy Fain \par

\cssec{\textsc{}Iteration 1}
\textless think\textgreater \par
Okay, let's tackle this step by step. The user is asking which American composer provided music for the musical 'Something More!'. 
First, I need to break down the question into sub-questions.\par
...\par
 if the results are clear, the answer can be extracted. If not, further refinements might be needed. But given the current state, the first search is essential to proceed.
 \par

\textless /think\textgreater \par

FUNCTION: retrievesearch \par
ARGS: \{"query": "American composer for Something More!", "top\_k": 5\} \par

\cssec{Memory update}
**Updated memory:**  
- The retrieved and recurrent chunks do not provide information about an American composer for "Something More!".  
- The band Syren is mentioned, but their album title is "Something Has To Separate," not "Something

\cssec{\textsc{}Iteration 2}
\textless think\textgreater \par
Okay, let's see. The user is asking which American composer provided music for the musical 'Something More!'.  \par
...\par

 but since the user's question is about the musical "Something More!", the next step is to search for the band Syren and their album title to see if the composer is part of that.\par
...\par

\textless /think\textgreater \par

FUNCTION: retrievesearch \par
ARGS: \{"query": "Syren Something Has To Separate composer", "top\_k": 5\} \par

\cssec{Memory update}
 The provided retrieved and recurrent chunks do not explicitly mention an American composer associated with the musical "Something More!". The closest reference is to the band Syren and their album **"Something Has To Separate"**, but this title is distinct from "Something More!". No composer is directly linked to "Something More!" in the given documents. 

**Updated memory:**  
- The retrieved and recurrent chunks do not provide information about an American composer for "Something More!".  
- The band Syren is mentioned, but their album title is "Something Has To Separate," not "Something More!".  
- No composer is explicitly linked to "Something More!" in the provided documents.  

The answer may not be present in the current retrieved or recurrent chunks, but further context or clarification on the title discrepancy (e.g., a typo or alternate title) might be needed to identify the correct composer.

\cssec{\textsc{}Iteration 3}
\textless think\textgreater \par
 Okay, let's see. The user is asking which American composer provided music for the musical 'Something More!'. The retrieval history shows that they've already done four steps of searching. \par
...\par
The current memory doesn't have that info, so the next step is to retrieve again with a query that includes "Something Has To Separate" and "composer" to get the necessary information.\par

\textless /think\textgreater \par

FUNCTION: retrievesearch \par
ARGS: \{"query": "Something Has To Separate composer", "top\_k": 5\} \par

\cssec{Memory update}
The retrieved chunk (Document 109) explicitly mentions that the musical **"Something More!"** has music composed by **Sammy Fain** (with lyrics by Marilyn Bergman and Alan Bergman) and additional contributions from Composer Robert Prince. This directly answers the problem. 

**Updated memory:**  
- The musical "Something More!" is composed by **Sammy Fain**, as stated in Document 109.  
- The band Syren's album "Something Has To Separate" is a distinct work and unrelated to "Something More!".  
- Composer Robert Prince also contributed to the musical's dance numbers.  

The answer is **Sammy Fain**.
\cssec{\textsc{PreThink}: early-stop decision}
\textless think\textgreater \par
 Okay, let's see. The user is asking which American composer provided music for the musical 'Something More!'.  \par
...\par
which is a different work. So the answer is already in the memory. No need for another retrieval. The answer is Sammy Fain.

\textless /think\textgreater \par

STOP\par

\cssec{Answer}
Sammy Fain \par
\end{bigcaseblue}

\caption{\textbf{Case study: early stopping enabled by \textsc{PreThink}.}
The agent uses explicit planning to decide whether to retrieve or stop.
It first issues broad queries that fail due to a title mismatch (confusing the musical \emph{Something More!} with an unrelated album),
then refines the query and retrieves decisive evidence that the musical’s music is by \textbf{Sammy Fain}.
Once the required fact is present in memory, \textsc{PreThink} triggers \emph{STOP} to avoid redundant searches and unnecessary memory overwrites.}

\label{fig:case_early_stop}
\end{figure*}

%% file: tables/InfMem_chunk_size_ablation.tex
\begin{table}[t]
\centering
\small
\setlength{\tabcolsep}{4pt}
\renewcommand{\arraystretch}{1.12}
\caption{\textbf{Fixed-budget comparison across retrieval chunk sizes.}
Under a constant retrieval budget, we vary the retrieved chunk size and report accuracy on long-context QA (HQA\_28k, SQD\_32k, MSQ\_32k, 2WK\_32k) and LongBench QA (Avg\_LB with per-task scores).
Overall, chunk size 500 achieves the best Avg\_RULER and Avg\_LB, suggesting a favorable balance between retrieval granularity and content diversity.}
\resizebox{\textwidth}{!}{%
\begin{tabular}{l| c|cccc |c|c cccc}
\toprule
\textbf{Setting} &
\cellcolor[HTML]{F3F3F3}\textbf{Avg\_RULER} &
\textbf{HQA\_28k} & \textbf{SQD\_32k} & \textbf{MSQ\_32k} & \textbf{2WK\_32k} &
\cellcolor[HTML]{F3F3F3}\textbf{Avg\_LB} &
\textbf{NQA} & \textbf{HQA} & \textbf{2Wiki} & \textbf{Qasper} & \textbf{MuSiQue} \\
\midrule
\texttt{chunk\_250\_top12} &
53.96 & 55.77 & \textbf{61.27} & 38.40 & 60.41 &
45.72 & 21.09 & 60.63 & 68.62 & 34.79 & 43.48 \\
\texttt{chunk\_500\_top6} &
\textbf{55.15} & \textbf{60.45} & 60.61 & \textbf{38.74} & \textbf{60.80} &
\textbf{48.27} & \textbf{24.19} & \textbf{61.56} & 68.95 & 35.29 & \textbf{51.37} \\
\texttt{chunk\_1000\_top3} &
53.65 & 58.52 & 61.05 & 35.75 & 59.26 &
46.37 & 22.19 & 57.84 & 71.97 & 35.01 & 44.84 \\
\texttt{chunk\_3000\_top1} &
44.59 & 50.37 & 47.43 & 30.92 & 49.64 &
45.68 & 19.98 & 60.19 & \textbf{72.82} & 36.14 & 39.29 \\
\midrule
\texttt{chunk\_2000\_top2} &
49.46 & 51.92 & 54.05 & 37.77 & 54.08 &
47.47 & 22.68 & 60.97 & 72.12 & \textbf{38.86} & 42.71 \\
\texttt{chunk\_1000\_top4} &
49.98 & 54.41 & 55.07 & 34.04 & 56.39 &
45.71 & 22.09 & 58.50 & 70.21 & 34.44 & 43.30 \\
\bottomrule
\end{tabular}
}

\label{tab:chunk_size_budget}
\end{table}

%% file: appendix/ablation.tex
\section{Ablation}
\label{app:ablation}

\subsection{Retrieval Chunk Size Selection}
\label{app:chunk_size}

We study the effect of retrieval chunk size under a fixed retrieval budget in Table~\ref{tab:chunk_size_budget}.
Specifically, we constrain the total retrieved context to approximately 3k tokens and vary the chunk size and corresponding top-$k$:
\texttt{chunk=250, top-$k=12$}, \texttt{chunk=500, top-$k=6$}, \texttt{chunk=1000, top-$k=3$}, \texttt{chunk=2000, top-$k=2$}, and \texttt{chunk=3000, top-$k=1$}.
Table~\ref{tab:chunk_size_budget} reports accuracy on long-context QA benchmarks (HQA 28k, SQuAD 32k, MuSiQue 32k, 2Wiki 32k) as well as LongBench QA.

Overall, a chunk size of 500 tokens achieves the best or near-best performance across most tasks.
Very small chunks (e.g., 250 tokens) provide fine-grained retrieval but can fragment semantically coherent evidence, increasing the burden on memory composition and cross-chunk reasoning.
Conversely, large chunks (e.g., 2000--3000 tokens) preserve local coherence but reduce content diversity under a fixed budget, increasing the risk that irrelevant context dilutes the decisive evidence.
The intermediate setting (\texttt{chunk=500, top-$k=6$}) strikes a favorable balance between retrieval granularity and evidence coverage, enabling InfMem to capture complementary facts while maintaining sufficient local context for reliable extraction and memory writing.

We additionally test a larger retrieval budget of approximately 4k tokens by including \texttt{chunk=1000, top-$k=4$} and \texttt{chunk=2000, top-$k=2$}. The same trend persists: the 500-token regime remains a robust sweet spot, suggesting that the optimal chunk size is primarily governed by the trade-off between granularity and diversity rather than the exact budget.

\subsection{Early Stop \& Latency Analysis}
\label{appendix:early_stop}

To investigate the real-world efficiency and performance of the InfMem framework, we evaluate our stopping policies on a subset of 128 test samples across the benchmarks. We deploy the models using vLLM to measure the actual serving latency alongside accuracy. Table \ref{tab:time_quality} summarizes the time and quality metrics corresponding to the visualization in Figure \ref{fig:4b_trend}. The results show that while the 1-stop variant offers the lowest latency, it suffers from performance degradation due to the premature truncation of evidence chains. In contrast, our default 3-stop variant occupies the Pareto frontier. It balances negligible computational overhead with significantly higher answer accuracy, confirming that a slightly conservative stopping policy preserves critical evidence without sacrificing practical serving efficiency.

\input{tables/table_for_preserved}

\begin{figure*}[t]
  \centering
  \begin{minipage}[t]{0.49\textwidth}
    \vspace{0pt} 
    \centering
    \adjustbox{valign=t}{\input{tables/think_or_not}}
    \captionof{table}{\textbf{Thinking-mode ablation for reproducing MemAgent-RL on Qwen3-4B.}
We compare Qwen3-4B with \emph{thinking mode} enabled vs.\ disabled when reproducing the MemAgent-RL pipeline.
Results show that activating thinking changes the agent’s tool-use behavior and leads to different LongBench QA outcomes.
Bold denotes the best score within this block, and underline denotes the runner-up.}
    \label{tab:longbench_think_nothink}
  \end{minipage}
  \hfill
  \begin{minipage}[t]{0.49\textwidth}
    \vspace{0pt} 
    \centering
    \includegraphics[width=\linewidth]{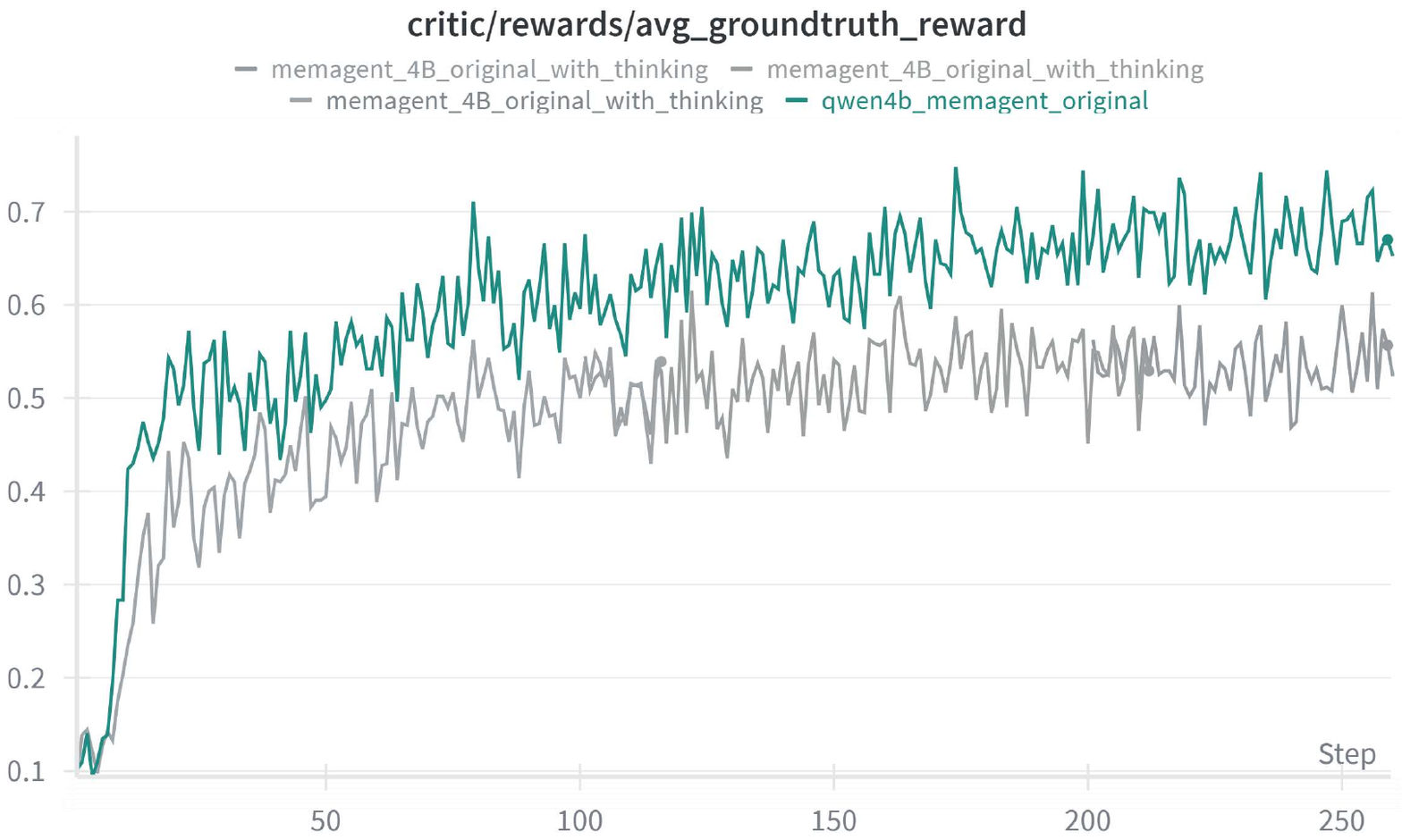}
    \caption{\textbf{Training dynamics: thinking vs.\ no-thinking for MemAgent-RL reproduction.}
We plot the training curves of reproduced MemAgent-RL runs on Qwen3-4B with thinking mode enabled and disabled. During training, the no-thinking variant consistently outperforms the thinking variant.}
    \label{fig:train_log}
  \end{minipage}
\end{figure*}

To further characterize this operational efficiency, Table \ref{tab:step_statistics} provides a detailed breakdown of the step distribution. Our method significantly reduces the total number of iteration steps through a highly adaptive and intelligent early stopping mechanism. InfMem autonomously modulates its search depth based on task complexity. For single-hop tasks like SQuAD, the framework rapidly converges on direct evidence, resulting in the shortest average step count. Conversely, for complex multi-hop tasks like MuSiQue, it deliberately traverses the document pool to resolve interdependent sub-queries. The early stop decision acts as a smart exit strategy: it terminates in very few steps when key evidence is quickly identified, yet retains the capacity to recurse deeply when initial searches are suboptimal, thereby optimizing execution speed without compromising the integrity of the evidence chain.

Table \ref{tab:latency_table} demonstrates the average total time proportion of different components per sample. Although our controller introduces new decision-making \& searching steps, building the BM25 index and the actual retrieval process consume only a marginal fraction of the total time. The substantial early stopping mechanism drastically cuts down the overall average latency and the time spent updating memory. Furthermore, as analyzed in subsequent sections, our framework generates higher-quality memory with less repetition; consequently, the time required to generate the final answer is also correspondingly reduced. Interestingly, when monitoring the theoretical average latency per sample, the speedup ratio over baselines is even more pronounced than the actual serving speedup. This discrepancy arises from the highly efficient continuous batching and queueing mechanisms of vLLM, which naturally optimize baselines like MemAgent that process a fixed number of chunks, whereas InfMem's actual batch throughput is occasionally dragged down by a few outlier samples that require maximum exploration steps without early stopping.

\input{tables/inference_time}

\label{app:think_or_not}

\input{tables/latency_table}
\input{figs/think_or_not_case_study}
\section{Ablation and Analysis of Thinking Dynamics}

\subsection{Rationale for Defaulting to No-Thinking in Baseline}
\label{app:ablation_baseline}

When reproducing MemAgent-RL on the Qwen3-series, we adopt the \textit{no-thinking} setting as the default configuration. This choice is primarily driven by \textbf{alignment with the original pipeline}, as the official MemAgent-RL setup (based on Qwen2.5-Instruct) does not inherently support thinking mode. However, beyond consistency, we empirically validate that enabling thinking in the baseline architecture is often counterproductive.

\input{figs/memory_quality}
The baseline MemAgent relies on a lightweight controller for naive compression, deciding strictly whether to write or skip the current chunk. In this regime, enabling thinking mode triggers an ``over-deliberation'' behavior. As illustrated in the case study (Figure~\ref{fig:comparison}), the agent expends substantial reasoning on loosely related chunks, which blurs the write/skip boundary. 

\textbf{Quantitative Evidence of Instability.} This destabilization is quantitatively captured in Table~\ref{tab:main_results}. While enabling thinking in the baseline (\textit{MemAgent Think-RL}) significantly improves the model's ability to discover answers (Found: 76.04\% vs. 72.46\%), it introduces severe volatility. The active reasoning process makes the memory vulnerable to recurrent noise, causing the \textit{Preserved} rate to drop sharply from 69.53\% to 65.43\%. This confirms that in the baseline architecture, the benefits of enhanced extraction are negated by the instability of memory updates, justifying our choice of the \textit{no-thinking} configuration for controlled comparisons.

\subsection{Decoupling Reasoning from Instability: The InfMem Advantage}
\label{app:quality}
Crucially, the analysis above raises a fundamental question: \textit{Is reasoning inherently detrimental to memory retention in recurrent systems?} Our results with \textbf{InfMem} suggest the answer is no. The instability observed in the baseline stems not from the thinking process itself, but from the \textbf{naive compression} mechanism that fails to filter the generated reasoning paths.

\textbf{Stability via Dynamic Chunking.} 
As shown in Table~\ref{tab:main_results}, \textsc{InfMem} effectively mitigates the extraction-retention trade-off. Despite leveraging reasoning to enhance information processing, \textsc{InfMem} maintains a robust retention profile. Its average Preserved rate (69.77\%) is not only significantly higher than the unstable \textit{MemAgent Think-RL} but is fully comparable to the conservative \textit{MemAgent NoThink-RL} baseline (69.53\%). This indicates that \textsc{InfMem}'s Dynamic Chunking successfully concentrates deliberation on salient regions, allowing the model to benefit from deep reasoning without succumbing to the forgetting issues typical of recurrent updates.

\textbf{Memory Purity and Downstream Performance.}
The advantages of \textsc{InfMem} extend beyond mere retention statistics to the \textbf{quality} of the preserved information (Fig~\ref{fig:memo_quality}). A key observation from Table~\ref{tab:main_results} is the performance discrepancy: while \textsc{InfMem} and \textit{MemAgent NoThink-RL} preserve a similar number of answers ($\sim$69\%), \textsc{InfMem} achieves substantially higher downstream performance (64.85\% vs. 56.94\%).

%% file: tables/table_for_preserved.tex
\begin{table*}[t]
\centering
\caption{Comparative Analysis of Memory Dynamics and Downstream Performance. The table reports Found rate, Preserved rate, and overall Performance across four datasets with varying context lengths.}
\label{tab:main_results}
\resizebox{\textwidth}{!}{%
\begin{tabular}{ll c ccc ccc ccc ccc}
\toprule
 & & \textbf{Avg} & \multicolumn{3}{c}{\textbf{HQA}} & \multicolumn{3}{c}{\textbf{SQD}} & \multicolumn{3}{c}{\textbf{MSQ}} & \multicolumn{3}{c}{\textbf{2WK}} \\
\cmidrule(lr){4-6} \cmidrule(lr){7-9} \cmidrule(lr){10-12} \cmidrule(lr){13-15}
\textbf{Model} & \textbf{Metric} & - & 28k & 56k & 112k & 32k & 64k & 128k & 32k & 64k & 128k & 32k & 64k & 128k \\
\midrule

\multirow{2}{*}{\textbf{\shortstack{MemAgent\\Think-RL}}} 
 & Found & 76.04 & 81.25 & 79.69 & 78.12 & 89.84 & 92.97 & 91.41 & 56.25 & 52.34 & 57.81 & 79.69 & 75.78 & 77.34 \\
 & Preserved & 65.43 & 72.66 & 67.97 & 61.72 & 78.12 & 85.16 & 78.12 & 47.66 & 44.53 & 42.97 & 73.44 & 67.19 & 65.62 \\
\midrule

\multirow{3}{*}{\textbf{\shortstack{MemAgent\\NoThink-RL}}} 
 & Found & 72.46 & 81.25 & 76.56 & 74.22 & 90.62 & 95.31 & 91.41 & 63.28 & 49.22 & 50.78 & 63.28 & 64.06 & 69.53 \\
 & Preserved & 69.53 & 78.12 & 74.22 & 72.66 & 89.84 & 95.31 & 91.41 & 60.16 & 44.53 & 46.88 & 57.03 & 60.94 & 63.28 \\
 & Performance & 56.94 & 71.18 & 66.21 & 62.42 & 69.49 & 69.84 & 72.96 & 41.79 & 41.55 & 36.62 & 56.43 & 48.55 & 46.18 \\
\midrule

\multirow{3}{*}{\textbf{InfMem}} 
 & Found & 74.61 & 75.78 & 71.88 & 75.00 & 78.91 & 78.91 & 78.91 & 60.16 & 64.06 & 61.72 & 85.16 & 82.81 & 82.03 \\
 & Preserved & 69.77 & 72.66 & 69.75 & 72.16 & 71.09 & 72.66 & 72.66 & 57.03 & 59.38 & 59.38 & 76.56 & 76.56 & 77.34 \\
 & Performance & 64.85 & 70.03 & 69.34 & 71.36 & 65.44 & 62.23 & 68.11 & 52.77 & 56.59& 55.59 & 70.55 & 67.22 & 68.91 \\

\bottomrule
\end{tabular}%
}
\label{tab:preserved}
\end{table*}

%% file: tables/think_or_not.tex
\centering
\scriptsize
\setlength{\tabcolsep}{4pt}
\renewcommand{\arraystretch}{1.15}
\resizebox{\textwidth}{!}{%
\begin{tabular}{l|cc|>{\columncolor{gray!15}}ccccc}
\toprule
 & MemAgent (No-think) & MemAgent (Think) & InfMem \\
\midrule
avg
 &  \underline{59.72} & 58.99 & \textbf{66.00} \\
 
\midrule
HQA\_28k
 &  \underline{71.18} & 59.20 & \textbf{71.44} \\

SQD\_32k
 &  \textbf{69.49} & 61.70 & \underline{65.31} \\

MSQ\_32k
 & 41.79 &  \underline{46.27} & \textbf{56.58} \\

2WK\_32k
 & 56.43 &  \underline{68.78} & \textbf{70.66} \\

\midrule
avg\_LB
 &  \underline{47.11} & 46.46 & \textbf{50.93} \\
\midrule
LB NQA
 &  \underline{20.74} & 20.43 & \textbf{20.77} \\

LB HQA
 & \underline{63.80} & 60.34 & \textbf{65.14} \\

LB 2Wiki
 &  \underline{67.83} & 65.19 & \textbf{74.76} \\

LB Qasper
 &  \underline{41.02} & 35.68 & \textbf{40.74} \\

LB Musique
 & 42.14 &  \underline{50.66} & \textbf{53.22} \\
\bottomrule
\end{tabular}
}

%% file: tables/inference_time.tex
\begin{table*}[t]
\centering
\small
\caption{\textbf{Effect of early-stopping strategies on performance and wall-clock time.}
We compare MemAgent (baseline) with two early-stop variants (1-stop and 3-stop) across three backbones.
Columns report Avg, HotpotQA (HQA), SQuAD, MuSiQue, and 2WikiMultihopQA (2Wiki), with \emph{Perf.} shown on the first row and \emph{Time} on the second row for each model.}
\setlength{\tabcolsep}{3.0pt}
\renewcommand{\arraystretch}{1.12}
\resizebox{\textwidth}{!}{%
\begin{tabular}{ll|r|rrrr|r|rrrr|r|rrrr}
\toprule
\multirow{2}{*}{\textbf{Model}} & \multirow{2}{*}{\textbf{Metric}} &
\multicolumn{5}{c|}{\textbf{MemAgent}} &
\multicolumn{5}{c|}{\textbf{InfMem 3-stop}} &
\multicolumn{5}{c}{\textbf{InfMem 1-stop}} \\
\cmidrule(lr){3-7}\cmidrule(lr){8-12}\cmidrule(lr){13-17}
&& \cellcolor[HTML]{F3F3F3}\textbf{Avg} & \textbf{HQA} & \textbf{SQuAD} & \textbf{MuSiQue} & \textbf{2Wiki}
& \cellcolor[HTML]{F3F3F3}\textbf{Avg} & \textbf{HQA} & \textbf{SQuAD} & \textbf{MuSiQue} & \textbf{2Wiki}
& \cellcolor[HTML]{F3F3F3}\textbf{Avg} & \textbf{HQA} & \textbf{SQuAD} & \textbf{MuSiQue} & \textbf{2Wiki} \\
\midrule

\multirow{2}{*}{\textbf{7B}}  & \textbf{Perf.} &
52.13 & 58.39 & 68.63 & 38.34 & 43.18 &
\textbf{63.00}& 57.51 & 67.71 & 58.57 & 68.20 &
\underline{59.86}& 54.01 & 70.10 & 51.56 & 63.76 \\
& \textbf{Time} &
51{:}34 & 43{:}48 & 50{:}09 & 54{:}37 & 57{:}44 &
\underline{21{:}35}& 28{:}10 & 19{:}00 & 18{:}19 & 20{:}49 &
\textbf{15{:}46}& 21{:}16 & 13{:}27 & 14{:}13 & 14{:}08 \\
\midrule

\multirow{2}{*}{\textbf{1.7B}} & \textbf{Perf.} &
36.59 & 42.50 & 47.29 & 24.05 & 32.52 &
\textbf{49.35}& 51.31 & 59.56 & 38.18 & 48.34 &
\underline{48.39}& 54.52 & 53.39 & 36.19 & 49.45 \\

& \textbf{Time} &
41{:}51 & 37{:}45 & 41{:}06 & 43{:}16 & 45{:}18 &
\underline{20{:}50}& 16{:}33 & 18{:}20 & 28{:}41 & 19{:}46 &
\textbf{12{:}41}& 11{:}03 & 10{:}52 & 16{:}20 & 12{:}28 \\
\midrule

\multirow{2}{*}{\textbf{4B}} & \textbf{Perf.} &
50.13 & 51.70 & 77.74 & 35.91 & 35.18 &
\textbf{65.80}& 66.13 & 73.81 & 56.86 & 66.39 &
\underline{61.80}& 62.91 & 66.19 & 50.45 & 67.65 \\

& \textbf{Time} &
60{:}45 & 51{:}31 & 64{:}09 & 59{:}37 & 67{:}44 &
\underline{23{:}59}& 27{:}33 & 18{:}40 & 19{:}00 & 30{:}42 &
\textbf{11{:}49}& 15{:}19 & 9{:}29 & 12{:}41 & 9{:}45 \\
\bottomrule
\end{tabular}
}

\label{tab:time_quality}
\end{table*}

%% file: tables/latency_table.tex
\begin{table}[htbp]
\centering
\caption{Average total time (s) across different context lengths. Speedup multipliers relative to \texttt{memagent} are shown as subscripts (\textcolor{teal}{teal} for speedup $>1$, \textcolor{red}{red} for speedup $<1$). Components are abbreviated: \textbf{Tot.} = Total, \textbf{B} = bm25\_build, \textbf{J/P} = judge\_or\_plan, \textbf{R} = retrieval, \textbf{MU} = memory\_update, \textbf{FA} = final\_answer.}
\label{tab:latency_table}
\resizebox{\textwidth}{!}{
\begin{tabular}{@{} l | c c c | c c c c c c | c c c c c c @{}}
\toprule
\multirow{2}{*}{\textbf{Length}} & \multicolumn{3}{c|}{\textbf{memagent}} & \multicolumn{6}{c|}{\textbf{infmem 3 stop}} & \multicolumn{6}{c}{\textbf{infmem 1 stop}} \\
\cmidrule(lr){2-4} \cmidrule(lr){5-10} \cmidrule(lr){11-16}
& \textbf{Tot.} & \textbf{MU} & \textbf{FA} 
& \textbf{Tot.} & \textbf{B} & \textbf{J/P} & \textbf{R} & \textbf{MU} & \textbf{FA} 
& \textbf{Tot.} & \textbf{B} & \textbf{J/P} & \textbf{R} & \textbf{MU} & \textbf{FA} \\
\midrule

\textbf{28k} 
& \textbf{32.79} & 32.52 & 0.27 
& \textbf{45.21$_{\textcolor{red}{\times 0.73}}$} & 0.019 & 23.70 & 0.002 & 21.16$_{\textcolor{teal}{\times 1.54}}$ & 0.33$_{\textcolor{red}{\times 0.81}}$ 
& \textbf{21.75$_{\textcolor{teal}{\times 1.51}}$} & 0.019 & 12.96 & 0.002 & 8.32$_{\textcolor{teal}{\times 3.91}}$ & 0.45$_{\textcolor{red}{\times 0.60}}$ \\

\textbf{56k} 
& \textbf{71.86} & 71.52 & 0.35 
& \textbf{51.77$_{\textcolor{teal}{\times 1.39}}$} & 0.037 & 28.62 & 0.003 & 22.81$_{\textcolor{teal}{\times 3.14}}$ & 0.30$_{\textcolor{teal}{\times 1.14}}$ 
& \textbf{36.60$_{\textcolor{teal}{\times 1.96}}$} & 0.037 & 22.48 & 0.003 & 13.83$_{\textcolor{teal}{\times 5.17}}$ & 0.25$_{\textcolor{teal}{\times 1.38}}$ \\

\textbf{112k} 
& \textbf{147.81} & 147.31 & 0.51 
& \textbf{65.08$_{\textcolor{teal}{\times 2.27}}$} & 0.076 & 39.64 & 0.006 & 25.16$_{\textcolor{teal}{\times 5.86}}$ & 0.21$_{\textcolor{teal}{\times 2.44}}$ 
& \textbf{44.66$_{\textcolor{teal}{\times 3.31}}$} & 0.076 & 30.47 & 0.004 & 13.87$_{\textcolor{teal}{\times 10.62}}$ & 0.24$_{\textcolor{teal}{\times 2.16}}$ \\

\textbf{224k} 
& \textbf{331.35} & 330.82 & 0.52 
& \textbf{93.48$_{\textcolor{teal}{\times 3.54}}$} & 0.156 & 61.69 & 0.010 & 31.39$_{\textcolor{teal}{\times 10.54}}$ & 0.23$_{\textcolor{teal}{\times 2.24}}$ 
& \textbf{72.95$_{\textcolor{teal}{\times 4.54}}$} & 0.156 & 53.24 & 0.009 & 19.27$_{\textcolor{teal}{\times 17.16}}$ & 0.27$_{\textcolor{teal}{\times 1.91}}$ \\

\textbf{448k} 
& \textbf{691.64} & 691.13 & 0.51 
& \textbf{148.57$_{\textcolor{teal}{\times 4.66}}$} & 0.321 & 109.92 & 0.025 & 38.08$_{\textcolor{teal}{\times 18.15}}$ & 0.22$_{\textcolor{teal}{\times 2.28}}$ 
& \textbf{123.60$_{\textcolor{teal}{\times 5.60}}$} & 0.321 & 99.31 & 0.022 & 23.69$_{\textcolor{teal}{\times 29.17}}$ & 0.25$_{\textcolor{teal}{\times 2.05}}$ \\

\textbf{896k} 
& \textbf{1499.21} & 1498.60 & 0.61 
& \textbf{311.08$_{\textcolor{teal}{\times 4.82}}$} & 0.676 & 225.41 & 0.109 & 84.45$_{\textcolor{teal}{\times 17.74}}$ & 0.43$_{\textcolor{teal}{\times 1.42}}$ 
& \textbf{229.50$_{\textcolor{teal}{\times 6.53}}$} & 0.676 & 190.34 & 0.055 & 38.08$_{\textcolor{teal}{\times 39.36}}$ & 0.35$_{\textcolor{teal}{\times 1.73}}$ \\

\bottomrule
\end{tabular}
}
\end{table}

%% file: figs/think_or_not_case_study.tex
\begin{figure*}[t]
\centering

\begin{tcbraster}[
  raster columns=2,
  raster column skip=8pt,
  raster equal height,
  raster valign=stretch
]

\begin{bigcase}{MemAgent (Static Chunking: Redundant reasoning on irrelevant context)}
\textless think\textgreater  \par
\textit{Okay, let's tackle this problem. The question is asking which film Huh Jung directed that was released on August 17, 2017. .. Since there's no conflicting information in the provided section, the previous memory remains accurate. So, the answer should still be "The Mimic."} \par
\textless /think\textgreater  \par
\cssec{Key Details from the Section}
\begin{itemize}[leftmargin=*]
    \item \textbf{Document 154} confirms \textit{Hide and Seek} (2013) as Huh Jung’s film, but this does not conflict with the previous memory.
    \item \textbf{No other films directed by Huh Jung in 2017} are mentioned in the provided section.
    \item \textbf{All other documents} in the section focus on unrelated topics, reaffirming the accuracy of the previous memory.
\end{itemize}

\textbf{Updated Memory:} Huh Jung directed "The Mimic," a horror film released on August 17, 2017. 
\end{bigcase}
\begin{bigcaseblue}{InfMem (Dynamic Chunking: Targeted evidence composition and efficient memory update)}
\textless think\textgreater  \par
\textit{Okay, let's tackle this problem step by step. The question is asking which film directed by Huh Jung was released on August 17, 2017 …. The note in the memory also confirms that there's no conflicting info, so the conclusion is valid.} \par
\textless /think\textgreater  \par
\cssec{Key evidence from the retrieved chunks}
\begin{itemize}[leftmargin=*]
    \item \textbf{Document 7 (retrieved chunk):} Confirms Huh Jung's second feature film is \textit{The Mimic}, which is "scheduled to release in 2017." The memory explicitly states the release date as August 17, 2017.
\end{itemize}

\cssec{Final Result}
The film directed by Huh Jung released on August 17, 2017, is \textbf{"The Mimic"}. \par
\textbf{Final note:} The retrieved chunks confirm Huh Jung's film and its release date, supporting the existing memory. \par
\textbf{Updated memory:} Huh Jung directed the film \textit{The Mimic}, which was released on August 17, 2017.
\end{bigcaseblue}

\end{tcbraster}
\caption{\textbf{Comparison of memory update efficiency between MemAgent and InfMem.} This snapshot is extracted from training logs at step 100. Left: MemAgent suffers from significant token redundancy due to exhaustive reasoning over irrelevant chunks, leading to over-thinking. Right: Our \method\ employs a \textbf{Dynamic Chunking} strategy within the PreThink-Retrieve-Write protocol, allowing the model to concentrate its reasoning capacity on critical evidence and update long-term memory with higher precision and lower computational cost.}
\label{fig:comparison}
\end{figure*}

%% file: figs/memory_quality.tex
\begin{figure*}[t]
\centering
\begin{tcbraster}[
  raster columns=2,
  raster column skip=8pt,
  raster equal height,
  raster valign=stretch
]

\begin{bigcase}
    \textbf{Query:} Nick Garcia played soccer for which Major League Soccer team as a member of the Eastern Conference? \\
    \tcbline
    \textbf{MemAgent (NoThink-RL) Process:} \\
    From Document 81, it is stated that Toronto FC is a team in the Major League Soccer Eastern Conference. However, this information is not directly applicable to Nick Garcia.\\
    From Document 17, it is stated that the Columbus Crew is a team in the Major League Soccer Eastern Conference. However, this information is not directly applicable to Nick Garcia. \\
    From Document 10, it is stated that Occhiuto played for the Buffalo Stallions \dots \\
    From Document 17, it is stated that the Columbus Crew is a team in the Major League Soccer Eastern Conference. \textcolor{red}{[Repetition]} \\
    From Document 10, it is stated that Occhiuto played for the Buffalo Stallions of the Major Indoor Soccer League. However, this information is not relevant to the problem. \\
    \dots \textit{(Repeats "Columbus Crew" and "Buffalo Stallions" which affect information density)}
\end{bigcase}
\begin{bigcaseblue}
    \textbf{Query:} Nick Garcia played soccer for which Major League Soccer team as a member of the Eastern Conference? \\
    \tcbline
    \textbf{InfMem (Ours) Process:} \\
    \textit{\textless think\textgreater} \\
    Okay, let's tackle this problem step by step. \dots I don't see any other mentions of Nick Garcia's team in the other documents. The other documents talk about other teams like Columbus Crew SC \dots but not Nick Garcia. Therefore, the answer is Toronto FC. \\
    \textit{\textless /think\textgreater} \\
    \vspace{0.1cm}
    The retrieved chunk (Document 109) explicitly states that \textbf{Nick Garcia played for Toronto FC}. Additionally, the memory section confirms that \textbf{Toronto FC is an MLS team in the Eastern Conference} (Document 81). \\
    \vspace{0.1cm}
    \textbf{\textcolor{blue}{Updated Memory:}} \\
    - Nick Garcia played for \textbf{Toronto FC} in MLS. \\
    - \textbf{Toronto FC} is a member of the Eastern Conference. \\
    \tcbline
    \textbf{Final Answer:} Nick Garcia played soccer for \textbf{Toronto FC} in the Eastern Conference.
\end{bigcaseblue}
\end{tcbraster}
\caption{\textbf{Qualitative Comparison of Memory Purity.} \textbf{Left:} MemAgent (NoThink) is susceptible to recurrent noise, tending to process and accumulate irrelevant information (e.g., details about unrelated teams like Columbus Crew) which dilutes memory utility. \textbf{Right:} In contrast, InfMem utilizes the reasoning mechanism to actively filter out these distractors. By synthesizing only the critical evidence, InfMem maintains a memory of \textbf{significantly higher quality and purity}, ensuring that only high-fidelity facts relevant to the query are preserved.}
\label{fig:memo_quality}
\end{figure*}

%% file: appendix/extended_results.tex
\section{Extended Results}
\label{app:extended}

\subsection{Generality Across Model Families}
\label{app:cross_family}

All backbones in the main results belong to the Qwen family.
To test whether the control loop is tied to that family, we run the \emph{identical} InfMem inference protocol on \texttt{Gemma-4-E4B-IT}, an instruction-tuned model of comparable effective size from a different training pipeline, \emph{without any SFT or RL on that backbone}, and compare it against MemAgent on the same model.
The evaluation setup follows the main comparison (HotpotQA at 112k--896k and MuSiQue at 128k--1M, 128 samples per length).

InfMem improves over MemAgent on all eight (task $\times$ length) configurations, by $+9.1$ to $+24.1$ F1 (Table~\ref{tab:gemma_transfer}), and reproduces the two qualitative patterns of the Qwen results: the gain is larger on multi-hop than on single-hop questions, and it grows with context length.
Most directly relevant to the question of attribution, the average marginal lift on multi-hop MuSiQue is essentially the same on the two families ($+20.3$ on Gemma vs.\ $+20.6$ on Qwen3-4B), indicating that the benefit of the control loop is a property of the protocol rather than an artifact of Qwen-specific training.
On single-hop HotpotQA the lift is somewhat larger on Gemma ($+13.9$ vs.\ $+10.1$), as the RL-trained Qwen baseline is already stronger on this task and leaves less headroom.
The efficiency profile transfers as well, with up to $5.0\times$ lower per-sample latency at 896k (Table~\ref{tab:gemma_efficiency}).
We note that InfMem on the zero-shot Gemma backbone still trails InfMem on the RL-trained \texttt{Qwen3-4B} by roughly $3$--$5$ F1 in absolute terms.

\paragraph{Model-family dispatch.}
Supporting a non-Qwen backbone requires no change to the control loop, only to prompt rendering and response parsing.
We add a tokenizer-based dispatch layer that detects the backbone family and selects the corresponding format: Qwen models use the native function-call markers, whereas Gemma uses the standard chat template with tool declarations and the model's tool-call and thinking delimiters.
Because those delimiters are special tokens, decoding for the Gemma path must preserve special tokens; the Qwen path is unchanged.

\input{tables/gemma_transfer}
\input{tables/gemma_efficiency}

\subsection{Additional Retrieval-based Baselines}
\label{app:extra_baselines}

Table~\ref{tab:extra_baselines} compares InfMem against two further families of retrieval-based systems under a matched in-document corpus.
\paragraph{Search-R1.}
Search-R1 is an RL-trained \emph{retrieval-decision} policy, and is therefore the closest learned-control baseline to InfMem.
We use the officially released \texttt{qwen2.5-7b-em-ppo} checkpoint and keep its trained \texttt{<think>/<search>/<information>/<answer>} loop intact, replacing only its open-domain Wikipedia/dense retriever with the same in-document BM25 index used by InfMem and MemAgent, so that the comparison isolates the quality of the search policy rather than the retrieval corpus.
Search-R1 never places the long document in its context: it emits a query and receives the top-3 in-document chunks as \texttt{<information>}, so its effective context stays small and its scores are nearly flat across lengths.
At the same backbone scale, it trails InfMem by $5.6$ F1 on HotpotQA and $8.5$ F1 on MuSiQue (averaged over the four lengths), and it is also below our 4B InfMem ($53.9/46.4$ vs.\ $68.1/58.4$).
It does clearly beat the no-control MemAgent baseline on multi-hop questions ($46.4$ vs.\ $39.2$), confirming that a learned retrieval decision helps; the InfMem loop helps more.

\paragraph{Self-RAG.}
We evaluate Self-RAG in two forms.
The first, denoted \emph{prompted Self-RAG-style RAG}, is a train-free pipeline on \texttt{Qwen3.6-35B-A3B} (thinking mode) that uses the same in-document BM25 retriever as our other baselines, prompts the model to emit Self-RAG-style reflection signals (retrieval, relevance, support and utility), and ranks candidate answers with the weighted arg-max rule of the original paper.
Despite a much larger backbone, it stays $24$--$27$ F1 below our 4B InfMem at every length on HotpotQA, and the gap widens sharply on multi-hop MuSiQue ($13.5/11.3/10.4/8.2$ F1 from 128k to 1M).

That collapse is localized to retrieval rather than to generation.
The fraction of questions whose retrieved passages actually contain the gold answer drops from $46.1\%$ on HotpotQA to $14.8\%$ on MuSiQue, whereas the fraction whose reasoning trace mentions the answer barely moves ($58.6\%$ to $50.8\%$); consequently the model abstains outright (``not provided in the given documents'') on $63$--$66\%$ of MuSiQue questions.
Across both tasks the F1 of this baseline tracks its retrieval hit rate almost exactly ($46.5$ F1 at $46.1\%$ hit rate on HotpotQA, $13.5$ F1 at $14.8\%$ on MuSiQue), i.e.\ a single question-shaped BM25 query essentially decides the outcome.
This is precisely the failure mode that query reformulation prevents: Search-R1 is also retrieval-only and likewise never sees the full document, but because it re-issues queries across turns it loses only $11\%$ relative when moving from HotpotQA to MuSiQue, against $71\%$ for this Self-RAG-style pipeline.

The second is a Self-RAG generator we trained ourselves on \texttt{Qwen3-4B-Instruct}, following the official recipe (Alpaca-format prompts, the 15 reflection tokens with mean-initialized embeddings, paragraph-content loss masking, and the $\approx$150K official Self-RAG training instances), in order to rule out the possibility that the official checkpoints underperform only because they are built on older Llama-2 backbones.
On Self-RAG's in-domain single-hop setting, PopQA, evaluated with the official short-form script, our checkpoint reaches $0.60$ accuracy and exceeds the official \texttt{selfrag\_llama2\_7b} checkpoint ($0.54$), confirming that it is correctly trained and competitive in the domain Self-RAG targets.

On our long-context benchmark, however, all Self-RAG variants remain far below InfMem (Table~\ref{tab:selfrag_subem}); we report sub-EM here because Self-RAG variants often emit reflection-token-style outputs rather than the short answers expected by our evaluator.
We evaluated the official checkpoints under both inference algorithms of the original paper, short-form (single-step per-passage reranking) and long-form (multi-step beam search), and verified that they emit valid reflection tokens, so the gap is not a formatting artifact.
The explanation is a task--paradigm mismatch rather than model capacity or checkpoint staleness: Self-RAG's retrieve-then-critique paradigm assumes that a single retrieved passage is largely sufficient to answer, whereas our setting requires synthesis across passages.
The short-form algorithm only ranks answers produced from individual passages, and the long-form algorithm performs no query reformulation, so it typically fails to retrieve second-hop evidence and effectively collapses to single-passage answering.

\input{tables/extra_baselines}
\input{tables/selfrag_subem}

\subsection{Token, Latency and Storage Accounting}
\label{app:cost_accounting}

\paragraph{Decode tokens and latency across lengths.}
Table~\ref{tab:token_latency_full} reports per-sample decode tokens together with per-sample latency for InfMem and MemAgent across six context lengths.
Below 56k, the fixed \textsc{PreThink} overhead makes InfMem slightly more expensive in tokens, while its latency is already on par.
From 112k onward InfMem dominates on both axes, saving $37$--$75\%$ of the tokens and $56$--$79\%$ of the wall-clock time.

As shown in Table~\ref{tab:step_statistics}, while the total number of chunks averages 202.8, InfMem 1-stop processes only 20.5 chunks on average (approximately 10\% of the full context), and InfMem 3-stop requires just 32.6 steps (16\%). This demonstrates that our adaptive stopping policy effectively identifies when sufficient information has been gathered, avoiding unnecessary computation over the remaining context.

Latency is saved faster than tokens (at 896k, $0.21\times$ the wall-clock but $0.28\times$ the tokens): MemAgent's cost grows with the number of chunks it must stream, whereas InfMem keeps a constant-size working memory and issues only a few retrieval rounds, so its wall-clock grows about $10\times$ from 28k to 896k against about $46\times$ for MemAgent.

\paragraph{Retrieval and memory budget.}
Building and querying the index is a negligible fraction of the total cost (Table~\ref{tab:latency_table}): InfMem issues about four retrieval calls per sample on average, and the BM25 build and lookup together account for well under $1\%$ of per-sample latency even at 896k.
The working memory is bounded by construction, at a 5k-token recurrent chunk plus a 1k-token memory slot per step, and is therefore constant in the input length.

\paragraph{Index storage.}
Because retrieval is performed over a per-document index, we also report its storage footprint (Table~\ref{tab:bm25_storage}).
With a production-grade Pyserini/Lucene backend, the minimal index needed for retrieval is $0.25$--$0.35\times$ the size of the raw text, and the ratio \emph{decreases} with document length, so the per-token cost is bounded: the index of a 914k-token document is only $885$\,KB.
Storing positions, doc-vectors and raw fields raises this to about $1.8\times$ raw, still below the $2.7$--$2.9\times$ of the uncompressed \texttt{rank\_bm25} reference implementation used in our main experiments.

\input{tables/step_statistics}
\input{tables/token_latency_full}
\input{tables/bm25_storage}

%% file: tables/gemma_transfer.tex
\begin{table*}[t]
\centering
\small
\setlength{\tabcolsep}{4pt}
\renewcommand{\arraystretch}{1.12}
\caption{\textbf{Zero-shot transfer of the InfMem control loop to a non-Qwen backbone.}
We run the identical InfMem inference protocol (\textsc{PreThink--Retrieve--Write}, 3-stop, BM25) on \texttt{Gemma-4-E4B-IT} \emph{without any SFT or RL on that backbone}, and compare against MemAgent on the same backbone.
The last row reports the corresponding framework lift on the RL-trained \texttt{Qwen3-4B}, computed from Table~\ref{tab:generality_backbones}, for reference.
On multi-hop MuSiQue the marginal lift is nearly identical across the two families ($+20.3$ vs.\ $+20.6$), indicating that the gain is attributable to the control loop rather than to Qwen-specific training.}
\label{tab:gemma_transfer}
\resizebox{\textwidth}{!}{%
\begin{tabular}{ll|ccccc|ccccc}
\toprule
\multirow{2}{*}{\textbf{Backbone}} & \multirow{2}{*}{\textbf{Method}} &
\multicolumn{5}{c|}{\textbf{HotpotQA}} &
\multicolumn{5}{c}{\textbf{MuSiQue}} \\
\cmidrule(lr){3-7}\cmidrule(lr){8-12}
& & \textbf{112k} & \textbf{224k} & \textbf{448k} & \textbf{896k} & \cellcolor[HTML]{F3F3F3}\textbf{Avg}
  & \textbf{128k} & \textbf{256k} & \textbf{512k} & \textbf{1M} & \cellcolor[HTML]{F3F3F3}\textbf{Avg} \\
\midrule
\multirow{3}{*}{\shortstack[l]{\textbf{Gemma-4-E4B-IT}\\ (zero-shot)}}
& MemAgent & 57.81 & 51.84 & 50.96 & 45.47 & \cellcolor[HTML]{F3F3F3}51.52
           & 19.60 & 26.80 & 19.90 & 23.22 & \cellcolor[HTML]{F3F3F3}22.38 \\
& \textbf{InfMem} & \textbf{66.91} & \textbf{64.41} & \textbf{67.08} & \textbf{63.20} & \cellcolor[HTML]{F3F3F3}\textbf{65.40}
                  & \textbf{39.45} & \textbf{41.16} & \textbf{43.99} & \textbf{45.98} & \cellcolor[HTML]{F3F3F3}\textbf{42.65} \\
\rowcolor[HTML]{FAFAFA}
& $\Delta$ & +9.10 & +12.57 & +16.12 & +17.73 & \cellcolor[HTML]{F3F3F3}\textbf{+13.88}
           & +19.85 & +14.36 & +24.09 & +22.76 & \cellcolor[HTML]{F3F3F3}\textbf{+20.27} \\
\midrule
\textbf{Qwen3-4B} (RL) & $\Delta$ (reference) & +8.82 & +8.30 & +8.91 & +14.43 & \cellcolor[HTML]{F3F3F3}+10.12
                       & +19.00 & +18.35 & +23.95 & +20.95 & \cellcolor[HTML]{F3F3F3}\textbf{+20.56} \\
\bottomrule
\end{tabular}}
\end{table*}

%% file: tables/gemma_efficiency.tex
\begin{table}[t]
\centering
\small
\setlength{\tabcolsep}{5pt}
\renewcommand{\arraystretch}{1.12}
\caption{\textbf{Inference efficiency of the zero-shot transfer to \texttt{Gemma-4-E4B-IT}.}
Left: per-sample latency (s) on HotpotQA. Right: total wall-clock (mm:ss, 128 samples, \texttt{n\_proc}=64) on MuSiQue, together with InfMem's \emph{found-in-memory} rate.
The efficiency profile matches the one observed on Qwen3-4B: the advantage grows with context length.}
\label{tab:gemma_efficiency}
\begin{tabular}{l|ccc|ccc}
\toprule
& \multicolumn{3}{c|}{\textbf{HotpotQA} (latency/sample, s)} & \multicolumn{3}{c}{\textbf{MuSiQue} (wall / found)} \\
\cmidrule(lr){2-4}\cmidrule(lr){5-7}
\textbf{Ctx} & \textbf{MemAgent} & \textbf{InfMem} & \textbf{Speedup}
& \textbf{MemAgent} & \textbf{InfMem} & \textbf{Found} \\
\midrule
112k / 128k & \phantom{0}109 & \phantom{0}63 & 1.7$\times$ & \phantom{0}5{:}31 & \phantom{0}6{:}17 & 66.41 \\
224k / 256k & \phantom{0}225 & \phantom{0}91 & 2.5$\times$ & 10{:}31 & 10{:}21 & 64.84 \\
448k / 512k & \phantom{0}593 & 134 & 4.4$\times$ & 22{:}20 & 18{:}55 & 65.62 \\
896k / 1M   & 1286 & 257 & \textbf{5.0$\times$} & 54{:}22 & 43{:}13 & 64.06 \\
\bottomrule
\end{tabular}
\end{table}

%% file: tables/extra_baselines.tex
\begin{table*}[t]
\centering
\small
\setlength{\tabcolsep}{4pt}
\renewcommand{\arraystretch}{1.12}
\caption{\textbf{Additional retrieval-based baselines under a matched in-document corpus.}
\textit{Search-R1} keeps its officially released RL-trained \texttt{<think>/<search>/<information>/<answer>} policy (\texttt{qwen2.5-7b-em-ppo}) and only swaps its open-domain Wikipedia/dense retriever for the same in-document BM25 used by InfMem and MemAgent, so that the comparison isolates the search policy rather than the retrieval corpus.
\textit{Self-RAG-style (prompted)} is a train-free Self-RAG-style reflection pipeline on a larger MoE backbone.
Bold denotes the best score per column within the 7B block.}
\label{tab:extra_baselines}
\resizebox{\textwidth}{!}{%
\begin{tabular}{ll|ccccc|ccccc}
\toprule
\multirow{2}{*}{\textbf{Backbone}} & \multirow{2}{*}{\textbf{Method}} &
\multicolumn{5}{c|}{\textbf{HotpotQA}} & \multicolumn{5}{c}{\textbf{MuSiQue}} \\
\cmidrule(lr){3-7}\cmidrule(lr){8-12}
& & \textbf{112k} & \textbf{224k} & \textbf{448k} & \textbf{896k} & \cellcolor[HTML]{F3F3F3}\textbf{Avg}
  & \textbf{128k} & \textbf{256k} & \textbf{512k} & \textbf{1M} & \cellcolor[HTML]{F3F3F3}\textbf{Avg} \\
\midrule
\multirow{3}{*}{\textbf{Qwen2.5-7B}}
& MemAgent & \textbf{61.55} & 59.95 & 57.09 & \textbf{58.39} & \cellcolor[HTML]{F3F3F3}59.2
           & 37.79 & 44.91 & 35.77 & 38.40 & \cellcolor[HTML]{F3F3F3}39.2 \\
& Search-R1 (RL) + in-doc BM25 & 52.61 & 54.95 & 53.87 & 54.01 & \cellcolor[HTML]{F3F3F3}53.9
           & 46.57 & 47.38 & 46.63 & 45.20 & \cellcolor[HTML]{F3F3F3}46.4 \\
\rowcolor{gray!15}
& \textbf{InfMem} & 57.75 & \textbf{60.55} & \textbf{62.34} & 57.51 & \cellcolor[HTML]{F3F3F3}\textbf{59.5}
           & \textbf{48.13} & \textbf{57.49} & \textbf{55.26} & \textbf{58.57} & \cellcolor[HTML]{F3F3F3}\textbf{54.9} \\
\midrule
\textbf{Qwen3.6-35B-A3B} & Self-RAG-style (prompted, train-free)
           & 46.48 & 43.05 & 41.26 & 42.13 & \cellcolor[HTML]{F3F3F3}43.2
           & 13.52 & 11.25 & 10.42 & \phantom{0}8.22 & \cellcolor[HTML]{F3F3F3}10.9 \\
\midrule
\rowcolor{gray!15}
\textbf{Qwen3-4B} & \textbf{InfMem (full, for reference)}
           & 71.24 & 67.42 & 67.75 & 66.13 & \cellcolor[HTML]{F3F3F3}68.1
           & 55.62 & 61.39 & 59.59 & 56.86 & \cellcolor[HTML]{F3F3F3}58.4 \\
\bottomrule
\end{tabular}}
\end{table*}

%% file: tables/selfrag_subem.tex
\begin{table}[t]
\centering
\small
\setlength{\tabcolsep}{6pt}
\renewcommand{\arraystretch}{1.12}
\caption{\textbf{Self-RAG variants on synthesized long-context HotpotQA (sub-EM, \%).}
We report sub-EM because Self-RAG variants frequently emit reflection-token-style outputs and do not consistently follow the short-answer instruction used by our evaluator.
Official checkpoints are evaluated under both inference algorithms of the original paper (short-form per-passage reranking and long-form beam search); we verified that they emit valid reflection tokens, so the low scores are not a formatting artifact.}
\label{tab:selfrag_subem}
\begin{tabular}{ll|rr}
\toprule
\textbf{Self-RAG variant} & \textbf{Backbone} & \textbf{800} & \textbf{6400} \\
\midrule
Official \texttt{selfrag\_llama2\_7b}  & Llama-2-7B (trained)               & 13.3 & 17.2 \\
Official \texttt{selfrag\_llama2\_13b} & Llama-2-13B (trained)              & \phantom{0}8.6 & 14.1 \\
Prompted Self-RAG-style RAG            & Qwen3.6-35B-A3B (train-free)       & 43.8 & 43.0 \\
Trained Self-RAG (ours)                & Qwen3-4B-Instruct (trained)        & 17.2 & 19.5 \\
\bottomrule
\end{tabular}
\end{table}

%% file: tables/step_statistics.tex
\begin{table}[h!]
\centering
\caption{Step-wise statistics of the InfMem variants across various benchmarks. The gap between average and maximum steps highlights the efficiency of the adaptive stopping policy.}
\label{tab:step_statistics}
\small
\begin{tabular}{@{}lrrrrr@{}}
\toprule
\rowcolor[HTML]{F3F3F3} 
\textbf{Metric} & \textbf{Avg. (All)} & \textbf{HQA} & \textbf{SQuAD} & \textbf{MuSiQue} & \textbf{2Wiki} \\ \midrule
Total Chunks & 202.8 & 181.1 & 209.8 & 210.0 & 210.1 \\ \midrule
\textbf{3-Stop Avg. Steps} & \textbf{32.6} & 27.9 & 20.1 & 51.1 & 31.0 \\
\quad Range (Min--Max) & 4--205 & 4--183 & 4--210 & 4--212 & 4--213 \\ \midrule
\textbf{1-Stop Avg. Steps} & \textbf{20.5} & 15.2 & 9.6 & 35.1 & 22.2 \\
\quad Range (Min--Max) & 2--204 & 24--183 & 2--210 & 2--211 & 2--212 \\ \bottomrule
\end{tabular}
\end{table}

%% file: tables/token_latency_full.tex
\begin{table}[t]
\centering
\small
\setlength{\tabcolsep}{4.5pt}
\renewcommand{\arraystretch}{1.12}
\caption{\textbf{Per-sample decode tokens and latency of InfMem (3-stop) vs.\ MemAgent across context lengths (Qwen3-4B, 128 samples per length, 4$\times$H100).}
Below 56k, the fixed \textsc{PreThink} overhead makes InfMem more expensive in tokens while latency is already on par; from 112k onwards InfMem dominates on both axes, and latency is saved faster than tokens, since MemAgent's per-step cost grows with the number of chunks whereas InfMem keeps a constant working memory plus a few retrieval rounds.}
\label{tab:token_latency_full}
\begin{tabular}{rr|rr|rr|rr}
\toprule
\multirow{2}{*}{\textbf{Len}} & \multirow{2}{*}{\textbf{Ctx}} &
\multicolumn{2}{c|}{\textbf{MemAgent}} & \multicolumn{2}{c|}{\textbf{InfMem}} &
\multicolumn{2}{c}{\textbf{Ratio}} \\
\cmidrule(lr){3-4}\cmidrule(lr){5-6}\cmidrule(lr){7-8}
& & \textbf{tok/Q} & \textbf{s/Q} & \textbf{tok/Q} & \textbf{s/Q} & \textbf{tok} & \textbf{s} \\
\midrule
\phantom{0}200 & $\sim$28k  & \phantom{00}2{,}192 & \phantom{00}32.8 & \phantom{0}4{,}422 & \phantom{0}45.3 & 2.02$\times$ & 1.38$\times$ \\
\phantom{0}400 & $\sim$56k  & \phantom{00}5{,}378 & \phantom{00}71.9 & \phantom{0}5{,}308 & \phantom{0}51.9 & 0.99$\times$ & 0.72$\times$ \\
\phantom{0}800 & $\sim$112k & \phantom{0}10{,}697 & \phantom{0}148.0 & \phantom{0}6{,}728 & \phantom{0}65.3 & \textbf{0.63$\times$} & \textbf{0.44$\times$} \\
1600 & $\sim$224k & \phantom{0}25{,}057 & \phantom{0}331.7 & \phantom{0}9{,}173 & \phantom{0}93.9 & \textbf{0.37$\times$} & \textbf{0.28$\times$} \\
3200 & $\sim$448k & \phantom{0}52{,}861 & \phantom{0}692.5 & 13{,}154 & 149.5 & \textbf{0.25$\times$} & \textbf{0.22$\times$} \\
6400 & $\sim$896k & 117{,}418 & 1501.0 & 33{,}085 & 312.9 & \textbf{0.28$\times$} & \textbf{0.21$\times$} \\
\bottomrule
\end{tabular}
\end{table}

%% file: tables/bm25_storage.tex
\begin{table}[t]
\centering
\small
\setlength{\tabcolsep}{4pt}
\renewcommand{\arraystretch}{1.12}
\caption{\textbf{Storage footprint of the in-document BM25 index} (chunk size 500, overlap 100; averaged over 16 samples per length).
\textit{Lucene min} is a production-grade Pyserini/Lucene index with only the fields needed for retrieval; \textit{Lucene full} additionally stores positions, doc-vectors and raw fields; \texttt{rank\_bm25} is the uncompressed pure-Python reference implementation used in our main experiments.
The index-to-raw ratio \emph{decreases} with document length, so the per-token storage cost is bounded: a 914k-token document needs only 885\,KB.}
\label{tab:bm25_storage}
\resizebox{\columnwidth}{!}{%
\begin{tabular}{r|rrr|rr|rr|rr}
\toprule
\multirow{2}{*}{\textbf{Len}} & \multirow{2}{*}{\textbf{\#Chunks}} & \multirow{2}{*}{\textbf{Ctx tok}} & \multirow{2}{*}{\textbf{Raw}} &
\multicolumn{2}{c|}{\textbf{Lucene min}} & \multicolumn{2}{c|}{\textbf{Lucene full}} & \multicolumn{2}{c}{\texttt{\textbf{rank\_bm25}}} \\
\cmidrule(lr){5-6}\cmidrule(lr){7-8}\cmidrule(lr){9-10}
& & & & \textbf{size} & \textbf{/raw} & \textbf{size} & \textbf{/raw} & \textbf{size} & \textbf{/raw} \\
\midrule
\phantom{0}800 & \phantom{0}281 & 112k & 442.6\,KB & \textbf{156\,KB} & \textbf{0.35$\times$} & 857\,KB & 1.94$\times$ & 1.3\,MB & 2.93$\times$ \\
1600 & \phantom{0}568 & 227k & 894\,KB   & \textbf{276\,KB} & \textbf{0.31$\times$} & 1.6\,MB & 1.88$\times$ & 2.5\,MB & 2.85$\times$ \\
3200 & 1140 & 456k & 1.7\,MB   & \textbf{489\,KB} & \textbf{0.27$\times$} & 3.2\,MB & 1.84$\times$ & 4.8\,MB & 2.78$\times$ \\
6400 & 2284 & 914k & 3.5\,MB   & \textbf{885\,KB} & \textbf{0.25$\times$} & 6.3\,MB & 1.80$\times$ & 9.5\,MB & 2.73$\times$ \\
\bottomrule
\end{tabular}}
\end{table}

%% file: colm2026_conference.bbl
\begin{thebibliography}{51}
\providecommand{\natexlab}[1]{#1}
\providecommand{\url}[1]{\texttt{#1}}
\expandafter\ifx\csname urlstyle\endcsname\relax
  \providecommand{\doi}[1]{doi: #1}\else
  \providecommand{\doi}{doi: \begingroup \urlstyle{rm}\Url}\fi

\bibitem[Achiam et~al.(2023)Achiam, Adler, Agarwal, Ahmad, Akkaya, Aleman,
  Almeida, Altenschmidt, Altman, Anadkat, et~al.]{achiam2023gpt}
Josh Achiam, Steven Adler, Sandhini Agarwal, Lama Ahmad, Ilge Akkaya,
  Florencia~Leoni Aleman, Diogo Almeida, Janko Altenschmidt, Sam Altman,
  Shyamal Anadkat, et~al.
\newblock Gpt-4 technical report.
\newblock \emph{arXiv preprint arXiv:2303.08774}, 2023.

\bibitem[An et~al.(2024)An, Gong, Zhong, Zhao, Li, Zhang, Kong, and
  Qiu]{an2024eval}
Chenxin An, Shansan Gong, Ming Zhong, Xingjian Zhao, Mukai Li, Jun Zhang,
  Lingpeng Kong, and Xipeng Qiu.
\newblock L-eval: Instituting standardized evaluation for long context language
  models.
\newblock In \emph{Proceedings of the 62nd Annual Meeting of the Association
  for Computational Linguistics (Volume 1: Long Papers)}, pp.\  14388--14411,
  2024.

\bibitem[Asai et~al.(2024)Asai, Wu, Wang, Sil, and Hajishirzi]{asai2024self}
Akari Asai, Zeqiu Wu, Yizhong Wang, Avirup Sil, and Hannaneh Hajishirzi.
\newblock Self-rag: Learning to retrieve, generate, and critique through
  self-reflection.
\newblock In \emph{The Twelfth International Conference on Learning
  Representations, {ICLR} 2024, Vienna, Austria, May 7-11, 2024}.
  OpenReview.net, 2024.
\newblock URL \url{https://openreview.net/forum?id=hSyW5go0v8}.

\bibitem[Bai et~al.(2024)Bai, Lv, Zhang, Lyu, Tang, Huang, Du, Liu, Zeng, Hou,
  Dong, Tang, and Li]{DBLP:conf/acl/BaiLZL0HDLZHDTL24}
Yushi Bai, Xin Lv, Jiajie Zhang, Hongchang Lyu, Jiankai Tang, Zhidian Huang,
  Zhengxiao Du, Xiao Liu, Aohan Zeng, Lei Hou, Yuxiao Dong, Jie Tang, and
  Juanzi Li.
\newblock Longbench: {A} bilingual, multitask benchmark for long context
  understanding.
\newblock In Lun{-}Wei Ku, Andre Martins, and Vivek Srikumar (eds.),
  \emph{Proceedings of the 62nd Annual Meeting of the Association for
  Computational Linguistics (Volume 1: Long Papers), {ACL} 2024, Bangkok,
  Thailand, August 11-16, 2024}, pp.\  3119--3137. Association for
  Computational Linguistics, 2024.
\newblock \doi{10.18653/V1/2024.ACL-LONG.172}.
\newblock URL \url{https://doi.org/10.18653/v1/2024.acl-long.172}.

\bibitem[Cao et~al.(2025)Cao, Cai, and Lam]{cao2025infiniteicl}
Bowen Cao, Deng Cai, and Wai Lam.
\newblock Infinite{ICL}: Breaking the limit of context window size via long
  short-term memory transformation.
\newblock In \emph{Findings of the Association for Computational Linguistics:
  ACL 2025}, 2025.

\bibitem[DeepSeek{-}AI(2025)]{deepseekai2025deepseekr1}
DeepSeek{-}AI.
\newblock Deepseek-r1: Incentivizing reasoning capability in llms via
  reinforcement learning.
\newblock \emph{CoRR}, abs/2501.12948, 2025.
\newblock \doi{10.48550/ARXIV.2501.12948}.
\newblock URL \url{https://doi.org/10.48550/arXiv.2501.12948}.

\bibitem[{DeepSeek-AI}(2025)]{deepseekdistillqwen_modelcard}
{DeepSeek-AI}.
\newblock Deepseek-r1-distill-qwen model card.
\newblock Hugging Face model repository, 2025.
\newblock URL
  \url{https://huggingface.co/deepseek-ai/DeepSeek-R1-Distill-Qwen-1.5B}.
\newblock Accessed 2026-01-28.

\bibitem[Gu \& Dao(2024)Gu and Dao]{gu2024mamba}
Albert Gu and Tri Dao.
\newblock Mamba: Linear-time sequence modeling with selective state spaces.
\newblock In \emph{First conference on language modeling}, 2024.

\bibitem[Ho et~al.(2020)Ho, Nguyen, Sugawara, and Aizawa]{2wiki}
Xanh Ho, Anh{-}Khoa~Duong Nguyen, Saku Sugawara, and Akiko Aizawa.
\newblock Constructing {A} multi-hop {QA} dataset for comprehensive evaluation
  of reasoning steps.
\newblock In Donia Scott, N{\'{u}}ria Bel, and Chengqing Zong (eds.),
  \emph{Proceedings of the 28th International Conference on Computational
  Linguistics, {COLING} 2020, Barcelona, Spain (Online), December 8-13, 2020},
  pp.\  6609--6625. International Committee on Computational Linguistics, 2020.
\newblock \doi{10.18653/V1/2020.COLING-MAIN.580}.
\newblock URL \url{https://doi.org/10.18653/v1/2020.coling-main.580}.

\bibitem[Jaech et~al.(2024)Jaech, Kalai, Lerer, Richardson, El-Kishky, Low,
  Helyar, Madry, Beutel, Carney, et~al.]{jaech2024openai}
Aaron Jaech, Adam Kalai, Adam Lerer, Adam Richardson, Ahmed El-Kishky, Aiden
  Low, Alec Helyar, Aleksander Madry, Alex Beutel, Alex Carney, et~al.
\newblock Openai o1 system card.
\newblock \emph{arXiv preprint arXiv:2412.16720}, 2024.

\bibitem[Jiang et~al.(2025)Jiang, Sun, Liang, and
  Zhang]{jiang2025retrievesummarizeplanadvancing}
Zhouyu Jiang, Mengshu Sun, Lei Liang, and Zhiqiang Zhang.
\newblock Retrieve, summarize, plan: Advancing multi-hop question answering
  with an iterative approach, 2025.
\newblock URL \url{https://arxiv.org/abs/2407.13101}.

\bibitem[Jin et~al.(2025)Jin, Zeng, Yue, Yoon, Arik, Wang, Zamani, and
  Han]{jin2025searchr1}
Bowen Jin, Hansi Zeng, Zhenrui Yue, Jinsung Yoon, Sercan Arik, Dong Wang, Hamed
  Zamani, and Jiawei Han.
\newblock Search-r1: Training llms to reason and leverage search engines with
  reinforcement learning.
\newblock \emph{arXiv preprint arXiv:2503.09516}, 2025.

\bibitem[Kahneman(2011)]{kahneman2011thinking}
Daniel Kahneman.
\newblock \emph{Thinking, fast and slow}.
\newblock macmillan, 2011.

\bibitem[Lewis et~al.(2020)Lewis, Perez, Piktus, Petroni, Karpukhin, Goyal,
  K{\"{u}}ttler, Lewis, Yih, Rockt{\"{a}}schel, Riedel, and
  Kiela]{DBLP:conf/nips/LewisPPPKGKLYR020}
Patrick Lewis, Ethan Perez, Aleksandra Piktus, Fabio Petroni, Vladimir
  Karpukhin, Naman Goyal, Heinrich K{\"{u}}ttler, Mike Lewis, Wen{-}tau Yih,
  Tim Rockt{\"{a}}schel, Sebastian Riedel, and Douwe Kiela.
\newblock Retrieval-augmented generation for knowledge-intensive {NLP} tasks.
\newblock In Hugo Larochelle, Marc'Aurelio Ranzato, Raia Hadsell,
  Maria{-}Florina Balcan, and Hsuan{-}Tien Lin (eds.), \emph{Advances in Neural
  Information Processing Systems 33: Annual Conference on Neural Information
  Processing Systems 2020, NeurIPS 2020, December 6-12, 2020, virtual}, 2020.
\newblock URL
  \url{https://proceedings.neurips.cc/paper/2020/hash/6b493230205f780e1bc26945df7481e5-Abstract.html}.

\bibitem[Li et~al.(2026)Li, Qi, Wu, Zhao, Ma, Li, Wang, Tai, Song, Zhang,
  Leung, and King]{li2026evipath}
Muzhi Li, Jinhu Qi, Yihong Wu, Minghao Zhao, Liheng Ma, Yifan Li, Xinyu Wang,
  Zhenghan Tai, Zixing Song, Yingxue Zhang, Ho-fung Leung, and Irwin King.
\newblock From evidence to trajectory: Abductive reasoning path synthesis for
  retrieval-augmented generation agents development.
\newblock In \emph{Proceedings of the 32nd ACM SIGKDD Conference on Knowledge
  Discovery and Data Mining (KDD)}, 2026.

\bibitem[Liu et~al.(2023)Liu, Zaharia, and Abbeel]{liu2023ring}
Hao Liu, Matei Zaharia, and Pieter Abbeel.
\newblock Ring attention with blockwise transformers for near-infinite context.
\newblock \emph{arXiv preprint arXiv:2310.01889}, 2023.

\bibitem[Liu et~al.(2024{\natexlab{a}})Liu, Lin, Hewitt, Paranjape, Bevilacqua,
  Petroni, and Liang]{liu2024lost}
Nelson~F Liu, Kevin Lin, John Hewitt, Ashwin Paranjape, Michele Bevilacqua,
  Fabio Petroni, and Percy Liang.
\newblock Lost in the middle: How language models use long contexts.
\newblock \emph{Transactions of the association for computational linguistics},
  12:\penalty0 157--173, 2024{\natexlab{a}}.

\bibitem[Liu et~al.(2024{\natexlab{b}})Liu, Lin, Hewitt, Paranjape, Bevilacqua,
  Petroni, and Liang]{lost_in_the_middle}
Nelson~F. Liu, Kevin Lin, John Hewitt, Ashwin Paranjape, Michele Bevilacqua,
  Fabio Petroni, and Percy Liang.
\newblock Lost in the middle: How language models use long contexts.
\newblock \emph{Trans. Assoc. Comput. Linguistics}, 12:\penalty0 157--173,
  2024{\natexlab{b}}.
\newblock \doi{10.1162/TACL\_A\_00638}.
\newblock URL \url{https://doi.org/10.1162/tacl\_a\_00638}.

\bibitem[Lu et~al.(2025)Lu, Huang, Zeng, Wang, Chen, Langlais, and
  Cui]{mamba_modulation}
Peng Lu, Jerry Huang, QIUHAO Zeng, Xinyu Wang, Boxing Chen, Philippe Langlais,
  and Yufei Cui.
\newblock Mamba modulation: On the length generalization of mamba models.
\newblock In \emph{The Thirty-ninth Annual Conference on Neural Information
  Processing Systems}, 2025.
\newblock URL \url{https://openreview.net/forum?id=QEU047bE8p}.

\bibitem[Ma et~al.(2026)Ma, Lo, Wang, Lu, Yuan, Chen, Han, Chen, Zhan, xu, Yin,
  Shang, Wen, Chen, and Cui]{match}
Linrui Ma, Chun~Hei Lo, Xinyu Wang, Peng Lu, Xihao Yuan, Hanting Chen, Kai Han,
  Xinghao Chen, Chengjun Zhan, Hanlin xu, Yichun Yin, Lifeng Shang, Feng Wen,
  Boxing Chen, and Yufei Cui.
\newblock {MATCH}: Modulating attention via in-context retrieval for
  long-context transformers.
\newblock In Maria Liakata, Viviane~P. Moreira, Jiajun Zhang, and David Jurgens
  (eds.), \emph{Proceedings of the 64th Annual Meeting of the {A}ssociation for
  {C}omputational {L}inguistics (Volume 1: Long Papers)}, pp.\  15165--15179,
  San Diego, California, United States, July 2026. Association for
  Computational Linguistics.
\newblock ISBN 979-8-89176-390-6.
\newblock \doi{10.18653/v1/2026.acl-long.692}.
\newblock URL \url{https://aclanthology.org/2026.acl-long.692/}.

\bibitem[Maharana et~al.(2024)Maharana, Lee, Tulyakov, Bansal, Barbieri, and
  Fang]{maharana2024evaluating}
Adyasha Maharana, Dong-Ho Lee, Sergey Tulyakov, Mohit Bansal, Francesco
  Barbieri, and Yuwei Fang.
\newblock Evaluating very long-term conversational memory of llm agents.
\newblock \emph{arXiv preprint arXiv:2402.17753}, 2024.

\bibitem[Packer et~al.(2023)Packer, Fang, Patil, Lin, Wooders, and
  Gonzalez]{packer2023memgpt}
Charles Packer, Vivian Fang, Shishir~G. Patil, Kevin Lin, Sarah Wooders, and
  Joseph~E. Gonzalez.
\newblock Memgpt: Towards llms as operating systems.
\newblock \emph{CoRR}, abs/2310.08560, 2023.
\newblock \doi{10.48550/ARXIV.2310.08560}.
\newblock URL \url{https://doi.org/10.48550/arXiv.2310.08560}.

\bibitem[Peng et~al.(2024{\natexlab{a}})Peng, Quesnelle, Fan, and
  Shippole]{peng2024yarn}
Bowen Peng, Jeffrey Quesnelle, Honglu Fan, and Enrico Shippole.
\newblock Yarn: Efficient context window extension of large language models.
\newblock In \emph{The Twelfth International Conference on Learning
  Representations, {ICLR} 2024, Vienna, Austria, May 7-11, 2024}.
  OpenReview.net, 2024{\natexlab{a}}.
\newblock URL \url{https://openreview.net/forum?id=wHBfxhZu1u}.

\bibitem[Peng et~al.(2024{\natexlab{b}})Peng, Quesnelle, Fan, and
  Shippole]{yarn}
Bowen Peng, Jeffrey Quesnelle, Honglu Fan, and Enrico Shippole.
\newblock Ya{RN}: Efficient context window extension of large language models.
\newblock In \emph{The Twelfth International Conference on Learning
  Representations}, 2024{\natexlab{b}}.
\newblock URL \url{https://openreview.net/forum?id=wHBfxhZu1u}.

\bibitem[Press et~al.(2021)Press, Smith, and Lewis]{press2021train}
Ofir Press, Noah~A Smith, and Mike Lewis.
\newblock Train short, test long: Attention with linear biases enables input
  length extrapolation.
\newblock \emph{arXiv preprint arXiv:2108.12409}, 2021.

\bibitem[Press et~al.(2022)Press, Smith, and Lewis]{train_short_test_long}
Ofir Press, Noah Smith, and Mike Lewis.
\newblock Train short, test long: Attention with linear biases enables input
  length extrapolation.
\newblock In \emph{International Conference on Learning Representations}, 2022.
\newblock URL \url{https://openreview.net/forum?id=R8sQPpGCv0}.

\bibitem[{Qwen Team}(2025)]{qwen2025qwen3next}
{Qwen Team}.
\newblock Qwen3-next: Hybrid attention and sparse moe (model release notes).
\newblock Qwen blog, 2025.
\newblock URL
  \url{https://qwen.ai/blog?from=research.latest-advancements-list&id=4074cca80393150c248e508aa62983f9cb7d27cd}.
\newblock Accessed 2026-01-28.

\bibitem[Rajpurkar et~al.(2016)Rajpurkar, Zhang, Lopyrev, and
  Liang]{squad_dataset}
Pranav Rajpurkar, Jian Zhang, Konstantin Lopyrev, and Percy Liang.
\newblock Squad: 100, 000+ questions for machine comprehension of text.
\newblock In Jian Su, Xavier Carreras, and Kevin Duh (eds.), \emph{Proceedings
  of the 2016 Conference on Empirical Methods in Natural Language Processing,
  {EMNLP} 2016, Austin, Texas, USA, November 1-4, 2016}, pp.\  2383--2392. The
  Association for Computational Linguistics, 2016.
\newblock \doi{10.18653/V1/D16-1264}.
\newblock URL \url{https://doi.org/10.18653/v1/d16-1264}.

\bibitem[Salvatore et~al.(2025)Salvatore, Wang, and
  Zhang]{lost_in_the_middle_ir}
Nikolaus Salvatore, Hao Wang, and Qiong Zhang.
\newblock Lost in the middle: An emergent property from information retrieval
  demands in llms.
\newblock \emph{CoRR}, abs/2510.10276, 2025.
\newblock \doi{10.48550/ARXIV.2510.10276}.
\newblock URL \url{https://doi.org/10.48550/arXiv.2510.10276}.

\bibitem[Shaham et~al.(2022)Shaham, Segal, Ivgi, Efrat, Yoran, Haviv, Gupta,
  Xiong, Geva, Berant, et~al.]{shaham2022scrolls}
Uri Shaham, Elad Segal, Maor Ivgi, Avia Efrat, Ori Yoran, Adi Haviv, Ankit
  Gupta, Wenhan Xiong, Mor Geva, Jonathan Berant, et~al.
\newblock Scrolls: Standardized comparison over long language sequences.
\newblock \emph{arXiv preprint arXiv:2201.03533}, 2022.

\bibitem[Su et~al.(2024)Su, Ahmed, Lu, Pan, Bo, and Liu]{su2024roformer}
Jianlin Su, Murtadha Ahmed, Yu~Lu, Shengfeng Pan, Wen Bo, and Yunfeng Liu.
\newblock Roformer: Enhanced transformer with rotary position embedding.
\newblock \emph{Neurocomputing}, 568:\penalty0 127063, 2024.

\bibitem[Sumers et~al.(2023)Sumers, Yao, Narasimhan, and
  Griffiths]{sumers2023cognitive}
Theodore Sumers, Shunyu Yao, Karthik~R Narasimhan, and Thomas~L Griffiths.
\newblock Cognitive architectures for language agents.
\newblock \emph{Transactions on Machine Learning Research}, 2023.

\bibitem[Touvron et~al.(2023)Touvron, Lavril, Izacard, Martinet, Lachaux,
  Lacroix, Rozi{\`e}re, Goyal, Hambro, Azhar, et~al.]{touvron2023llama}
Hugo Touvron, Thibaut Lavril, Gautier Izacard, Xavier Martinet, Marie-Anne
  Lachaux, Timoth{\'e}e Lacroix, Baptiste Rozi{\`e}re, Naman Goyal, Eric
  Hambro, Faisal Azhar, et~al.
\newblock Llama: Open and efficient foundation language models.
\newblock \emph{arXiv preprint arXiv:2302.13971}, 2023.

\bibitem[Trivedi et~al.(2022)Trivedi, Balasubramanian, Khot, and
  Sabharwal]{musique}
Harsh Trivedi, Niranjan Balasubramanian, Tushar Khot, and Ashish Sabharwal.
\newblock Musique: Multihop questions via single-hop question composition.
\newblock \emph{Trans. Assoc. Comput. Linguistics}, 10:\penalty0 539--554,
  2022.
\newblock \doi{10.1162/TACL\_A\_00475}.
\newblock URL \url{https://doi.org/10.1162/tacl\_a\_00475}.

\bibitem[Wan et~al.(2025{\natexlab{a}})Wan, Shen, Liao, Shi, Li, Yang, Zhang,
  Huang, Zhou, and Yan]{alibabacloud2025qwenlong}
Fanqi Wan, Weizhou Shen, Shengyi Liao, Yingcheng Shi, Chenliang Li, Ziyi Yang,
  Ji~Zhang, Fei Huang, Jingren Zhou, and Ming Yan.
\newblock Qwenlong-l1: Towards long-context large reasoning models with
  reinforcement learning.
\newblock \emph{CoRR}, abs/2505.17667, 2025{\natexlab{a}}.
\newblock \doi{10.48550/ARXIV.2505.17667}.
\newblock URL \url{https://doi.org/10.48550/arXiv.2505.17667}.

\bibitem[Wan et~al.(2025{\natexlab{b}})Wan, Shen, Liao, Shi, Li, Yang, Zhang,
  Huang, Zhou, and Yan]{wan2025qwenlongl1}
Fanqi Wan, Weizhou Shen, Shengyi Liao, Yingcheng Shi, Chenliang Li, Ziyi Yang,
  Ji~Zhang, Fei Huang, Jingren Zhou, and Ming Yan.
\newblock Qwenlong-l1: Towards long-context large reasoning models with
  reinforcement learning, 2025{\natexlab{b}}.
\newblock URL \url{https://arxiv.org/abs/2505.17667}.

\bibitem[Wang et~al.(2025{\natexlab{a}})Wang, Wang, Wang, Li, Tang, Hong,
  Chang, Wu, and Liu]{wang2025care}
Suyuchen Wang, Jinlin Wang, Xinyu Wang, Shiqi Li, Xiangru Tang, Sirui Hong,
  Xiao-Wen Chang, Chenglin Wu, and Bang Liu.
\newblock Improving context fidelity via native retrieval-augmented reasoning.
\newblock In \emph{Proceedings of the 2025 Conference on Empirical Methods in
  Natural Language Processing (EMNLP)}, 2025{\natexlab{a}}.

\bibitem[Wang et~al.(2025{\natexlab{b}})Wang, Ma, Huang, Lu, Parthasarathi,
  Chang, Chen, and Cui]{wang2025resona}
Xinyu Wang, Linrui Ma, Jerry Huang, Peng Lu, Prasanna Parthasarathi, Xiao-Wen
  Chang, Boxing Chen, and Yufei Cui.
\newblock Resona: Improving context copying in linear recurrence models with
  retrieval.
\newblock In \emph{Conference on Language Modeling (COLM)}, 2025{\natexlab{b}}.

\bibitem[Wang et~al.(2026)Wang, Lyu, Gu, Huang, Lu, Cui, and
  Chang]{wang2026rlsedit}
Xinyu Wang, Sicheng Lyu, Yu~Gu, Jerry Huang, Peng Lu, Yufei Cui, and Xiao-Wen
  Chang.
\newblock Beyond hard writes and rigid preservation: Soft recursive
  least-squares for lifelong {LLM} editing.
\newblock \emph{arXiv preprint arXiv:2601.15686}, 2026.

\bibitem[Wang et~al.(2025{\natexlab{c}})Wang, Takanobu, Liang, Mao, Hu,
  McAuley, and Wu]{wang2025mem}
Yu~Wang, Ryuichi Takanobu, Zhiqi Liang, Yuzhen Mao, Yuanzhe Hu, Julian McAuley,
  and Xiaojian Wu.
\newblock Mem-$\{$$\backslash$alpha$\}$: Learning memory construction via
  reinforcement learning.
\newblock \emph{arXiv preprint arXiv:2509.25911}, 2025{\natexlab{c}}.

\bibitem[Wei et~al.(2022)Wei, Wang, Schuurmans, Bosma, Xia, Chi, Le, Zhou,
  et~al.]{wei2022chain}
Jason Wei, Xuezhi Wang, Dale Schuurmans, Maarten Bosma, Fei Xia, Ed~Chi, Quoc~V
  Le, Denny Zhou, et~al.
\newblock Chain-of-thought prompting elicits reasoning in large language
  models.
\newblock \emph{Advances in neural information processing systems},
  35:\penalty0 24824--24837, 2022.

\bibitem[Weston \& Sukhbaatar(2023)Weston and Sukhbaatar]{weston2023system}
Jason Weston and Sainbayar Sukhbaatar.
\newblock System 2 attention (is something you might need too).
\newblock \emph{arXiv preprint arXiv:2311.11829}, 2023.

\bibitem[Yan et~al.(2025)Yan, Yang, Huang, Nie, Ding, Li, Ma, Kersting, Pan,
  Sch{\"u}tze, et~al.]{yan2025memory}
Sikuan Yan, Xiufeng Yang, Zuchao Huang, Ercong Nie, Zifeng Ding, Zonggen Li,
  Xiaowen Ma, Kristian Kersting, Jeff~Z Pan, Hinrich Sch{\"u}tze, et~al.
\newblock Memory-r1: Enhancing large language model agents to manage and
  utilize memories via reinforcement learning.
\newblock \emph{arXiv preprint arXiv:2508.19828}, 2025.

\bibitem[Yang et~al.(2024{\natexlab{a}})Yang, Yang, Zhang, Hui, Zheng, Yu, Li,
  Liu, Huang, Wei, Lin, Yang, Tu, Zhang, Yang, Yang, Zhou, Lin, Dang, Lu, Bao,
  Yang, Yu, Li, Xue, Zhang, Zhu, Men, Lin, Li, Xia, Ren, Ren, Fan, Su, Zhang,
  Wan, Liu, Cui, Zhang, and Qiu]{DBLP:journals/corr/abs-2412-15115}
An~Yang, Baosong Yang, Beichen Zhang, Binyuan Hui, Bo~Zheng, Bowen Yu,
  Chengyuan Li, Dayiheng Liu, Fei Huang, Haoran Wei, Huan Lin, Jian Yang,
  Jianhong Tu, Jianwei Zhang, Jianxin Yang, Jiaxi Yang, Jingren Zhou, Junyang
  Lin, Kai Dang, Keming Lu, Keqin Bao, Kexin Yang, Le~Yu, Mei Li, Mingfeng Xue,
  Pei Zhang, Qin Zhu, Rui Men, Runji Lin, Tianhao Li, Tingyu Xia, Xingzhang
  Ren, Xuancheng Ren, Yang Fan, Yang Su, Yichang Zhang, Yu~Wan, Yuqiong Liu,
  Zeyu Cui, Zhenru Zhang, and Zihan Qiu.
\newblock Qwen2.5 technical report.
\newblock \emph{CoRR}, abs/2412.15115, 2024{\natexlab{a}}.
\newblock \doi{10.48550/ARXIV.2412.15115}.
\newblock URL \url{https://doi.org/10.48550/arXiv.2412.15115}.

\bibitem[Yang et~al.(2025{\natexlab{a}})Yang, Li, Yang, Zhang, Hui, Zheng, Yu,
  Gao, Huang, Lv, Zheng, Liu, Zhou, Huang, Hu, Ge, Wei, Lin, Tang, Yang, Tu,
  Zhang, Yang, Yang, Zhou, Lin, Dang, Bao, Yang, Yu, Deng, Li, Xue, Li, Zhang,
  Wang, Zhu, Men, Gao, Liu, Luo, Li, Tang, Yin, Ren, Wang, Zhang, Ren, Fan, Su,
  Zhang, Zhang, Wan, Liu, Wang, Cui, Zhang, Zhou, and
  Qiu]{DBLP:journals/corr/abs-2505-09388}
An~Yang, Anfeng Li, Baosong Yang, Beichen Zhang, Binyuan Hui, Bo~Zheng, Bowen
  Yu, Chang Gao, Chengen Huang, Chenxu Lv, Chujie Zheng, Dayiheng Liu, Fan
  Zhou, Fei Huang, Feng Hu, Hao Ge, Haoran Wei, Huan Lin, Jialong Tang, Jian
  Yang, Jianhong Tu, Jianwei Zhang, Jian Yang, Jiaxi Yang, Jingren Zhou,
  Junyang Lin, Kai Dang, Keqin Bao, Kexin Yang, Le~Yu, Lianghao Deng, Mei Li,
  Mingfeng Xue, Mingze Li, Pei Zhang, Peng Wang, Qin Zhu, Rui Men, Ruize Gao,
  Shixuan Liu, Shuang Luo, Tianhao Li, Tianyi Tang, Wenbiao Yin, Xingzhang Ren,
  Xinyu Wang, Xinyu Zhang, Xuancheng Ren, Yang Fan, Yang Su, Yichang Zhang,
  Yinger Zhang, Yu~Wan, Yuqiong Liu, Zekun Wang, Zeyu Cui, Zhenru Zhang,
  Zhipeng Zhou, and Zihan Qiu.
\newblock Qwen3 technical report.
\newblock \emph{CoRR}, abs/2505.09388, 2025{\natexlab{a}}.
\newblock \doi{10.48550/ARXIV.2505.09388}.
\newblock URL \url{https://doi.org/10.48550/arXiv.2505.09388}.

\bibitem[Yang et~al.(2025{\natexlab{b}})Yang, Yu, Li,
  et~al.]{yang2025qwen25_1m}
An~Yang, Bowen Yu, Chengyuan Li, et~al.
\newblock Qwen2.5-1m technical report, 2025{\natexlab{b}}.
\newblock URL \url{https://arxiv.org/abs/2501.15383}.

\bibitem[Yang et~al.(2024{\natexlab{b}})Yang, Wang, Zhang, Shen, and
  Kim]{yang2024parallelizing}
Songlin Yang, Bailin Wang, Yu~Zhang, Yikang Shen, and Yoon Kim.
\newblock Parallelizing linear transformers with the delta rule over sequence
  length.
\newblock \emph{Advances in neural information processing systems},
  37:\penalty0 115491--115522, 2024{\natexlab{b}}.

\bibitem[Yang et~al.(2018)Yang, Qi, Zhang, Bengio, Cohen, Salakhutdinov, and
  Manning]{hotpot_qa}
Zhilin Yang, Peng Qi, Saizheng Zhang, Yoshua Bengio, William~W. Cohen, Ruslan
  Salakhutdinov, and Christopher~D. Manning.
\newblock Hotpotqa: {A} dataset for diverse, explainable multi-hop question
  answering.
\newblock In Ellen Riloff, David Chiang, Julia Hockenmaier, and Jun'ichi Tsujii
  (eds.), \emph{Proceedings of the 2018 Conference on Empirical Methods in
  Natural Language Processing, Brussels, Belgium, October 31 - November 4,
  2018}, pp.\  2369--2380. Association for Computational Linguistics, 2018.
\newblock \doi{10.18653/V1/D18-1259}.
\newblock URL \url{https://doi.org/10.18653/v1/d18-1259}.

\bibitem[Yu et~al.(2025)Yu, Chen, Feng, Chen, Dai, Yu, Zhang, Ma, Liu, Wang,
  and Zhou]{yu2025memagent}
Hongli Yu, Tinghong Chen, Jiangtao Feng, Jiangjie Chen, Weinan Dai, Qiying Yu,
  Ya-Qin Zhang, Wei-Ying Ma, Jingjing Liu, Mingxuan Wang, and Hao Zhou.
\newblock Memagent: Reshaping long-context llm with multi-conv rl-based memory
  agent, 2025.
\newblock URL \url{https://arxiv.org/abs/2507.02259}.

\bibitem[Zhang et~al.(2023)Zhang, Chen, Zhang, Xu, Zhao, and
  Yu]{zhang2023large}
Danyang Zhang, Lu~Chen, Situo Zhang, Hongshen Xu, Zihan Zhao, and Kai Yu.
\newblock Large language models are semi-parametric reinforcement learning
  agents.
\newblock \emph{Advances in Neural Information Processing Systems},
  36:\penalty0 78227--78239, 2023.

\bibitem[Zhou et~al.(2025)Zhou, Qu, Wu, Kim, Prakash, Rus, Zhao, Low, and
  Liang]{zhou2025mem1}
Zijian Zhou, Ao~Qu, Zhaoxuan Wu, Sunghwan Kim, Alok Prakash, Daniela Rus,
  Jinhua Zhao, Bryan Kian~Hsiang Low, and Paul~Pu Liang.
\newblock Mem1: Learning to synergize memory and reasoning for efficient
  long-horizon agents.
\newblock \emph{arXiv preprint arXiv:2506.15841}, 2025.

\end{thebibliography}
